\documentclass[acmsmall]{acmart}
\setcopyright{none}
\renewcommand\footnotetextcopyrightpermission[1]{}

\AtBeginDocument{%
  \providecommand\BibTeX{{%
    \normalfont B\kern-0.5em{\scshape i\kern-0.25em b}\kern-0.8em\TeX}}}

\usepackage{booktabs} 
\usepackage{graphicx}
\usepackage{balance}  
\usepackage{xspace}
\usepackage{hyperref}
\usepackage{url}
\usepackage{color}
\usepackage{xcolor}
\usepackage{fancyvrb}
\usepackage[center]{subfigure} 
\usepackage{multirow}
\usepackage{enumitem}
\usepackage{amsmath}
\usepackage{amsfonts}

\usepackage[ruled,vlined,linesnumbered]{algorithm2e}
\usepackage{algorithmic}

\newtheorem{problem}{Problem}
\newtheorem{definition}{Definition}

\newcommand{\method}{\textsc{Acorn}\xspace}

\newcommand{\fdata}{{SEC-seq}\xspace}

\newcommand{\T}[1]{\boldsymbol{\mathcal{#1}}}
\newcommand{\mat}[1]{\mathbf{#1}}

\newcommand{\footnoteref}[1]{\textsuperscript{\ref{#1}}}

\newcommand{\subfloat}{\subfigure}

\newcommand\green[1]{\textcolor{black}{#1}}

\begin{document}

\title{Accurate Open-set Recognition for Memory Workload}

\author{Jun-Gi Jang}
\email{elnino4@snu.ac.kr}
\affiliation{
	\institution{Seoul National University}
	\city{Seoul}
	\country{South Korea}
}

\author{Sooyeon Shim}
\email{syshim77@snu.ac.kr}
\affiliation{
  \institution{Seoul National University}
  \city{Seoul}
  \country{South Korea}
}

\author{Vladimir Egay}
\email{waldemar@snu.ac.kr}
\affiliation{
	\institution{Seoul National University}
	\city{Seoul}
	\country{South Korea}
}

\author{Jeeyong Lee}
\email{jiyong.lee@samsung.com}
\affiliation{
	\institution{Samsung Electronics}
	\country{South Korea}
}

\author{Jongmin Park}
\email{j1014.park@samsung.com}
\affiliation{
	\institution{Samsung Electronics}
	\country{South Korea}
}

\author{Suhyun Chae}
\email{suhyun.chae@samsung.com}
\affiliation{
	\institution{Samsung Electronics}
	\country{South Korea}
}

\author{U Kang}
\authornote{Corresponding author.
\authornotemark[1]}
\email{ukang@snu.ac.kr}
\affiliation{
  \institution{Seoul National University}
  \city{Seoul}
  \country{South Korea}
}

\renewcommand{\shortauthors}{Jang et al.}

\begin{abstract}
How can we accurately identify new memory workloads while classifying known memory workloads?
%
Verifying DRAM (Dynamic Random Access Memory) using various workloads is an important task
to guarantee the quality of DRAM.
A crucial component in the process is open-set recognition which aims to detect new workloads not seen in the training phase.
Despite its importance, however, existing open-set recognition methods are unsatisfactory in terms of accuracy since they fail to exploit the characteristics of workload sequences. 

In this paper, we propose \method, an accurate open-set recognition method capturing the characteristics of workload sequences.
\method extracts two types of feature vectors to capture sequential patterns and spatial locality patterns in memory access.
\method then uses the feature vectors to accurately classify a subsequence into one of the known classes or identify it as the unknown class.
Experiments show that \method achieves state-of-the-art accuracy,
giving up to $37\%$ points higher unknown class detection accuracy while achieving comparable known class classification accuracy than existing methods.
\end{abstract}



\maketitle

\section{Introduction}
\label{sec:introduction}
\textit{How can we accurately identify new memory workloads while classifying known workloads?}
The global DRAM (Dynamic Random Access Memory) market size is about tens of \green{billions} USD, and keeps increasing due to growing demand of DRAM in mobile devices, modern computers, self-driving cars, etc.
It is crucial to test DRAM using various workloads in verifying and guaranteeing DRAM quality.
DRAM manufacturers utilize their known workloads for verification; however, it does not guarantee that DRAM works well for new workloads not known in advance.
Therefore, it is necessary to detect new workloads to improve the quality of DRAM verification.
\green{The problem of detecting new workloads is formulated as an open-set recognition~\cite{ScheirerRSB13} task which classifies a test sample into the known classes or the unknown class, and identifies its class if it belongs to the known classes.}

A workload sequence contains a series of tuples with the command and the address information of memory accesses.
To detect new workloads based on \green{open-set recognition,} we exploit a subsequence, a part of the entire sequence of a workload.
Given a subsequence, we classify it into one of the known workload classes or identify it as the unknown class corresponding to new workloads.
Although there are several works~\cite{BendaleB16,ShuXL17,HassenC20,LiangLS18,LeeYY20} for the open-set recognition problem, none of them handles workload sequences.
Their accuracy is limited \green{for workload sequences} since
they do not exploit the characteristics of them.
The major challenges to be tackled are 1) how to deal with very long subsequences (e.g., $100,000$), 
and  2) how to detect subsequences generated from new workloads not seen in \green{the training phase.}

\green{
We provide an example of how DRAM manufacturers use an open-set recognition method to improve DRAM verification.
Executing a code generates a workload sequence which contains a series of tuples with the command and the address information of memory accesses.
Assume that there is a code that frequently provokes memory failures, and there is a situation where we do not have the code but only have its workload sequence.
Then, DRAM manufacturers want to design a test code similar to the failure-generating code since they need to verify DRAM for these failures.
%
If an accurate open-set recognition method exists, the manufacturers utilize it to compare their code with failure-generating code, and design a new test code that generates failures. 
In addition, if a given sequence belongs to the unknown workload class, we train a new classifier with existing classes and the new class of the given sequence; 
then, we can precisely classify even the workload for the unknown class. 
This process helps prevent DRAM failures and improve DRAM quality.
Therefore, we need to devise an accurate open-set recognition method for memory workload.
}

In this paper, we propose \method, an \textbf{\underline{AC}}curate \textbf{\underline{O}}pen-set recognition method for wo\textbf{\underline{R}}kload seque\textbf{\underline{N}}ces, to classify a subsequence of a workload into known classes or identify it as the unknown class that has not been observed during training.
To the best of our knowledge, \method is the first open-set recognition method for workload sequences.
\method obtains feature vectors of subsequences by exploiting the characteristics of workload sequences.
We split the workload fields into cmd and address-related fields, and extract features for each type.
For the cmd field, we exploit $n$-gram models to capture sequential patterns and construct a feature vector using frequent $n$-grams.
For the address-related fields, we construct a feature vector that captures the spatial locality patterns by counting the number of accesses to memory regions we carefully define.
This process makes the proposed method extract representative patterns from the workloads.
For the unknown class detection, we adopt the concept of anomaly detection based on a dimensionality reduction technique where abnormal test samples generate large reconstruction errors.
Our main idea is to build an unknown class detector for \textit{each class} using \textit{our feature vectors}, to detect the pattern of the unknown class which deviates significantly from those of the known classes.
It leads to accurate detection for test subsequences of the unknown class. 
%
%
Experimental results show that \method achieves the state-of-the-art performance in terms of both known class classification and unknown class detection, compared to baselines.

We summarize our main contributions as follows:


\begin{itemize}[noitemsep,topsep=0pt]
    \item \textbf{Problem formulation and data.}
    We formulate the new problem of open-set recognition for workload sequences (Problem~\ref{problem:openset_workload}), and release Memtest86-seq\footnote{\url{https://github.com/snudatalab/Acorn}},
     the first public dataset containing a large-scale memory workload sequence generated from open-domain programs\footnote{\url{https://www.memtest86.com/}}.
    \item \textbf{Method.} We propose \method, an effective and accurate method which extracts representative features from long workload subsequences and performs known class classification as well as unknown class detection.
    \item \textbf{Experiment.} \method outperforms existing open-set recognition methods by up to $37\%$ points higher unknown class detection accuracy with comparable known class accuracy than existing methods.
\end{itemize}


\green{
The rest of the paper is organized as follows:
we give the related works and the problem definition in Section \ref{sec:prelim}, propose \method in Section \ref{sec:method}, show the experimental results in Section \ref{sec:experiment}, and conclude in Section \ref{sec:conclusion}.
The code and the dataset are available at \textbf{\url{https://github.com/snudatalab/Acorn}}.
}

\section{Preliminaries}
\label{sec:prelim}
In this section, we describe \green{the preliminaries} and our problem definition.
The notations used in this paper are given in Table~\ref{tab:notation}.

\begin{table}[t]
	\centering
	\caption{Symbols used in this paper.}
	\label{tab:notation}
		\begin{tabular}{cl}
			\toprule
			\textbf{Symbol} & \textbf{Description} \\
			\midrule
			$\mathbf{W}$ & workload sequence matrix \\
			$\mathbf{S}_{i}$ & $i$-th subsequence matrix for training \\			
			$\mathcal{A}_n$ & set of distinct n-grams \\
			$|\T{A}_n|$ & the cardinality of a set $\T{A}_n$ \\
			$\mat{x}_{i, CMD}$ & the feature vector of cmd field for $i$-th training subsequence \\
			$\mat{x}_{i, ADDRESS}$ & the feature vector of address-related fields for $i$-th training subsequence \\
			$\mat{V}_{w}$ & unknown class detector matrix for a known class $w$ \\
			$\epsilon_{w}$ & the threshold for a known class $w$ \\
			\bottomrule
		\end{tabular}
\end{table}

\subsection{Workload Sequence}
\label{subsec:workload_sequence}

We use the term \textit{workload sequence} to define a sequence of commands produced by a DRAM controller unit during the whole process of program execution.
\begin{definition}[Workload Sequence]
	 A workload sequence $\mathbf{W} \in \mathbb{R}^{l\times 5}$ is a multidimensional sequence with the five fields cmd, rank, bank group, bank, and address, where $l$ is the length of the sequence.
%
	 \begin{itemize}
	 	\item Command (cmd) - can be one of the following 5 commands: ACT, RDA, WRA, PRE, and PREA.
	 	\item Rank - rank number inside a DRAM.
	 	\item Bank Group - bank group number within a rank.
	 	\item Bank - bank number within a bank group.
	 	\item Address - corresponds to a row or column address within a bank.
	 \end{itemize}
\end{definition}
Each row of a sequence consists of a command and $4$ address-related values.
A value of the cmd field represents a type of operation or command.
\green{
Address-related values for rank, bank group, bank, and address fields indicate the exact location in DRAM to which a command is applied.}

DRAM Controller produces $25$ commands, and there are $5$ representative commands in the cmd field: ACT, RDA, WRA, PRE, and PREA.
In a bank, ACT command is passed along with a row address to activate a row for the RDA or WRA command.
When a row is activated, RDA or WRA commands can be transmitted along with a column number indicating column Read or Write followed by precharge operation.
After all Read/Write operations to the activated row are completed, the PRE (precharge) command deactivates the row, so the next ACT command for a different row can be transmitted.
Note that we cannot activate more than two rows at the same time in a bank.
The PREA command deactivates all the active rows in all banks of a rank, so all the banks in the rank are ready to be accessed.
\green{We refer the reader to~\cite{ddr4} for further details about the cmd field.}

\begin{figure*}[!t]
	\centering
	\vspace{-2mm}
	\includegraphics[width=0.95\textwidth]{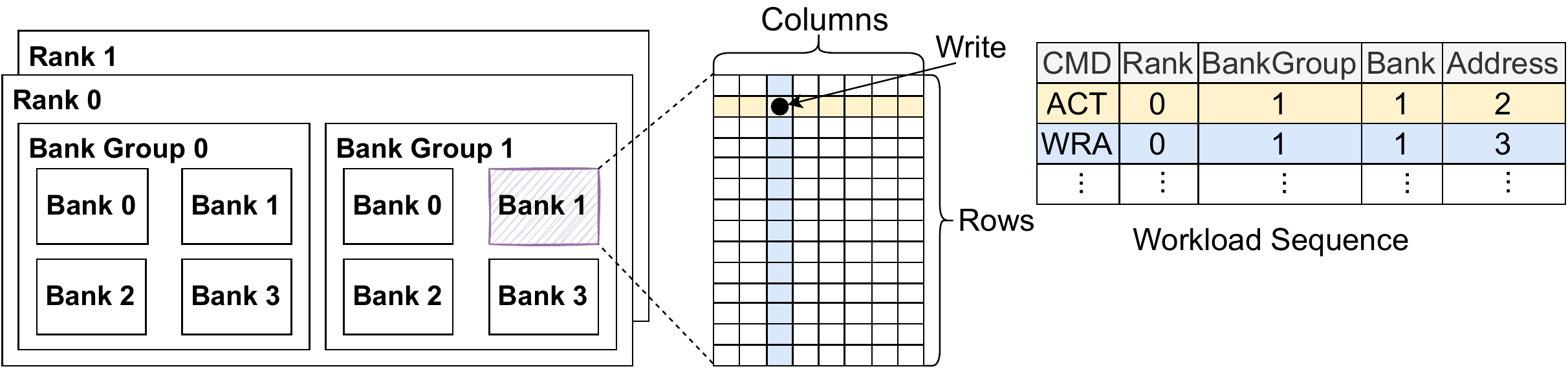}
	\caption{Workload sequence and address hierarchy in DRAM.}
	\label{fig:rank_address_hiearchy}
\end{figure*}

The values of address-related fields point out a specific location of a command operation.
The rank is the highest level of organization which consists of several bank groups, and the bank group is a collection of banks.
The value of the bank field is the index of a bank in the specified bank group.
We can track the exact bank, which is a two-dimensional array whose cells store data, by combining the values of the rank, bank group, and bank fields since DRAM has a hierarchical structure.
\green{The address field of a workload sequence contains the information on the bank's row number when the cmd field is ACT and a column number when the cmd field is either RDA or WRA.}
To detect the target memory cell where RDA/WRA command has been performed, one needs to find the preceding ACT command at the same rank, bank group, and bank.
For example, Fig.~\ref{fig:rank_address_hiearchy} shows that write operation at location (2, 3) of rank 0, bank group 1, bank 1 since row number 2 was previously activated by the preceding ACT command with the same bank location.


\subsection{Open-set Recognition}
\label{subsec:openset}

Open-set recognition aims to detect samples from the unknown class not included in the training dataset, while classifying samples from known classes.
\green{Open-set recognition has been utilized for many real-world applications and is more challenging than semi-supervised learning~\cite{YooJK19,ChenYZWCN20,LuoCNYHZ18} due to the unknown class detection.
Zero-shot learning~\cite{LiYCZSZ19,YooCKK19} also identifies the unknown class, but it requires external knowledge which helps differentiate known and unknown classes.}
Since none of the existing methods take memory workloads as input, we describe previous model-agnostic methods that can be applied to open-set recognition for workload sequences.
Model-agnostic methods can be combined with various deep learning-based models including MLP and CNN.
Bendale et al.~\cite{BendaleB16} propose OpenMax layer which extends softmax for open-set recognition.
Shu et al.~\cite{ShuXL17} apply open-set recognition to sequence domain, introducing a document open classification (DOC) model that utilizes 1-vs-rest layer with a sigmoid function as an alternative to a softmax layer.
Hassen et al.~\cite{HassenC20} introduce ii-loss which forces a network to maximize the distance between given classes and minimize the distance between an instance and the center of its class in the feature space.
Out-of-Distribution (OOD) methods~\cite{LiangLS18,LeeLLS18,LeeYY20,LiuWOL20,Sun21} can be also \green{applied to detect} new workloads.
Although the above methods have been applied to workload sequences, they fail to exploit the characteristics of the workload sequences.
Yoshihashi et al.~\cite{YoshihashiSKYIN19}, Oza et al.~\cite{OzaP19}, and Sun et al.~\cite{SunYZLP20} also address open-set recognition, but they require specific networks in contrast to the above methods.
\subsection{Problem Definition}
\label{subsec:problem_definition}


We use the term \textit{workload subsequence} to define a sub-part of the workload which has been cut to the same length.
\begin{definition}[Workload Subsequence]
	 Given a workload sequence matrix $\mat{W} \in \mathbb{R}^{l \times 5}$ where $l$ is the length of the sequence and $5$ is the number of the fields, a workload subsequence $\mathbf{S}^{(j)} \in \mathbb{R}^{l_s \times 5}$ is the $j$th vertical block matrix of $\mathbf{W}$ when $\mathbf{W}$ is vertically partitioned by length $l_s$ without overlapping: $\mat{W} = \begin{bmatrix} \mat{S}^{(1)} \\ \vdots \\ \mat{S}^{(l/l_s)} \end{bmatrix}$.
	 For simplicity, we represent a subsequence as $\mat{S}$ by dropping the notation $(j)$ describing the $j$th vertical block.
\end{definition}
In this paper, we set the length $l_s$ of subsequences to $100,000$.
We collect all subsequences $\mathbf{S}$ from all known workload sequences, and then randomly pick them to construct a set $\{(\mat{S}_{1}, y_1), ..., (\mat{S}_{N}, y_N)\}$ of training samples.
$N$ is the number of training samples, $\mat{S}_{i}$ is the $i$th training subsequence, and $y_i$ is the label that indicates the workload that generated $\mat{S}_i$.
The number of classes is equal to the number of known workloads when we train a model.
Test samples consist of all subsequences for all unknown workloads and the subsequences not picked as the training samples for all known workloads.

We introduce the formal problem definition as follows:
\begin{problem}[Workload Open-set Recognition]
\label{problem:openset_workload}
	\textbf{Given} a memory workload subsequence, \textbf{classify} it into one of the known classes or \textbf{identify} it as the unknown class.
	\begin{itemize}[noitemsep,topsep=0pt]
		\item Known workloads are seen in the training phase, and a known workload corresponds to its known class.
		\item Unknown workloads are not seen in the training phase, and all unknown workloads correspond to the unknown class.
	\end{itemize}
\end{problem}


\section{Proposed Method}
\label{sec:method}

In this section, we propose \method, an accurate open-set recognition method for workload sequences.
We need to tackle the following challenges:
\begin{itemize}[noitemsep,topsep=0pt]
	\item[C1.] \textbf{Dealing with heterogeneous fields.} How can we deal with $5$ heterogeneous fields, i.e., cmd, rank, bank group, bank, and address?
	\item[C2.] \textbf{Dealing with long subsequences for workload classification.} It is impractical to train a classification model using workload subsequences whose length is $0.1$ million.
		How can we deal with long subsequences?
	\item[C3.] \textbf{Detecting new workloads unseen at the training phase.}
How can we identify unseen workloads that appear only in the test phase?
\end{itemize}

\begin{figure*}[t]
	\centering
	\vspace{-2mm}
	\includegraphics[width=0.99\textwidth]{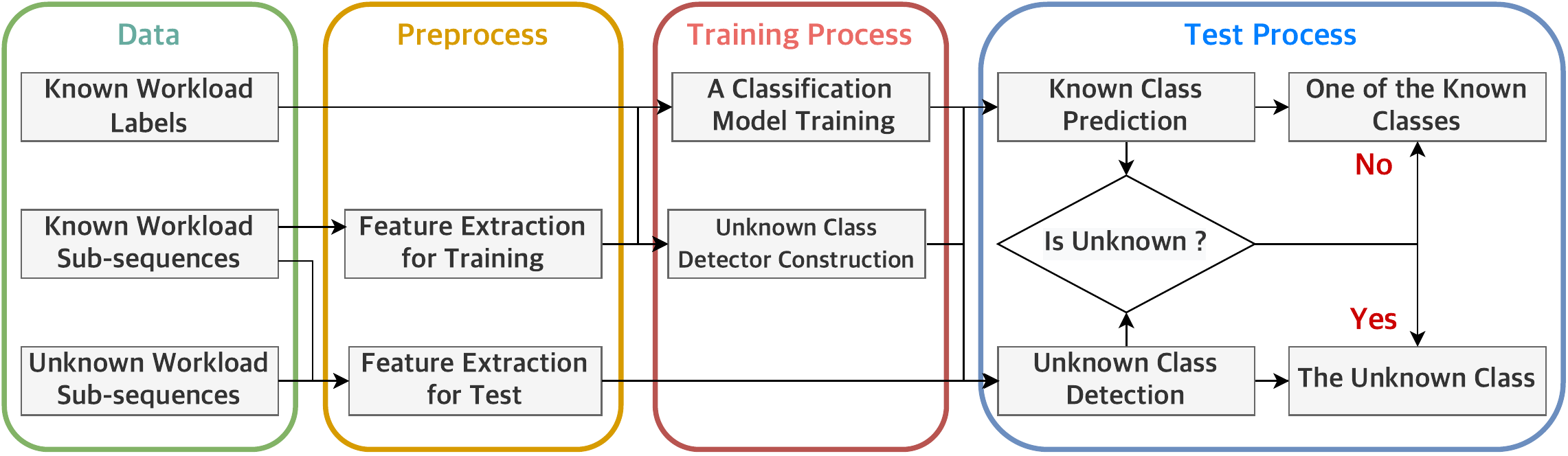}
	\caption{The overall process for \method on workload open-set recognition.}
	\label{fig:flowchart}
\end{figure*}

\noindent To achieve high accuracy for known class classification and unknown class detection, we propose the following main ideas:
\begin{itemize}[noitemsep,topsep=0pt]
	\item[I1.] \textbf{Discrimination-aware handling.} The values of the cmd field and the address-related fields indicate \green{the type} of the operation and the location of the operation, respectively. We process the fields by considering the difference.
	\item[I2.] \textbf{Capturing sequential patterns and spatial locality patterns.} We transform a long subsequence into two types of feature vectors of small sizes, and capture sequential and spatial patterns from the cmd and the address-related fields, respectively.
	\item[I3.] \textbf{Reconstruction error-based unknown class detection.}
	Constructing an unknown class detector for each known class makes test subsequences of the unknown class have high reconstruction errors, clearly distinguishing them from those of the known classes.
\end{itemize}

Fig.~\ref{fig:flowchart} shows the overall process for \method.
We construct training data with known workloads and test data using both known and unknown workloads.
In the training process, we extract features from subsequences.
Then, we train a classification model and construct unknown class detectors using the extracted features.
In the test process,
we find a feature vector of a test subsequence, predict a label $\hat{w}$ using the classification model, and then use a detector to identify whether it belongs to the label $\hat{w}$ or the unknown class.


\subsection{Feature Extraction}
\label{subsec:cons_vectors}
\green{The most important challenge is to effectively deal with a long subsequence with heterogeneous fields, while achieving high accuracy for workload classification.}
A naive approach is to train a classification model directly using subsequences.
However, a subsequence $\mat{S}_{i} \in \mathbb{R}^{100,000 \times 5}$ is too large to be used as an input for training a classification model.
Moreover, the values in each field have different meanings.
For example, $1$ in the cmd field indicates \green{an \textit{ACT} command} while $1$ in the bank field indicates \textit{the second bank} number in a bank group.
Therefore, we need to extract a valuable feature vector from a subsequence.
Our main ideas are to 1) separate the fields into two types, the cmd and the address-related fields (i.e., rank, bank group, bank, and address), and 2) consider the different characteristics of the two types in the fields.

\begin{figure*}[t]
	\centering
	\vspace{-2mm}
	\includegraphics[width=0.99\textwidth]{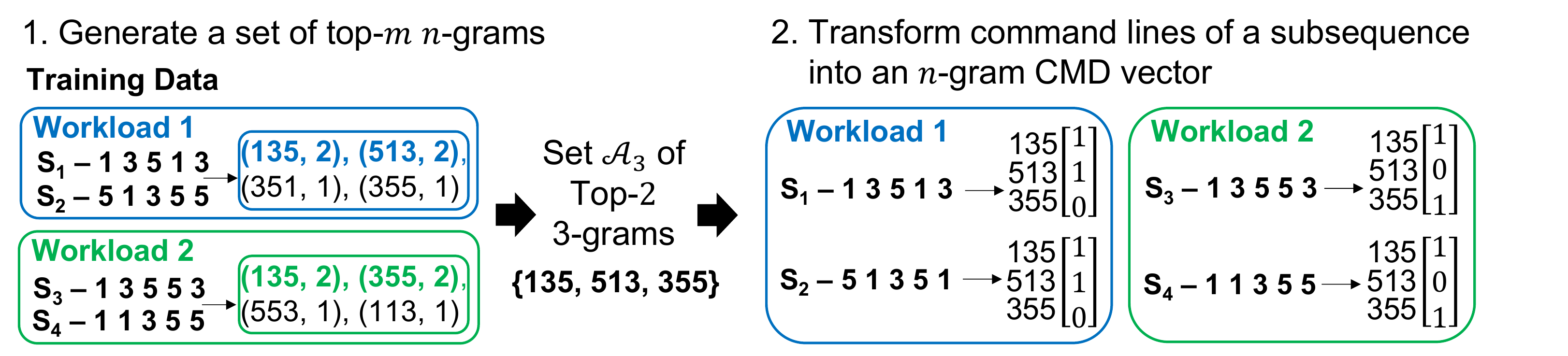}
	\caption{An example of generating a set of top-$m$ $n$-grams and an $n$-gram CMD vector of a subsequence. There are two workload sequences with $m=2$ and $n=3$.}
	\label{fig:example_cmd_vectors}
\end{figure*}

\textbf{CMD Feature Vector for Command Field.}
We first focus on transforming command lines $\mat{S}_{i}(:, 0) \in \mathbb{R}^{100,000 \times 1}$ of a subsequence into a feature vector of small size while capturing crucial information.
Since each workload has a different order of occurring commands, capturing sequential patterns in the command lines is important; hence we exploit $n$-gram models, used in various applications~\cite{TomovicJK06,ZhangXMNMS19,Ali20,Mutinda21}, which count a contiguous sequence of $n$ commands.
We construct a set $\T{A}_n$ of $n$-gram sequences that frequently appear in workloads, and then transform command lines $\mat{S}_{i}(:,0)$ of a subsequence into a feature vector $\mat{c}_{i, n} \in \mathbb{R}^{|\T{A}_n|}$ using the set $\T{A}_n$ where $|\T{A}_n|$ is the cardinality of the set $\T{A}_n$.
To be more specific, we count $n$-gram sequences from subsequences selected as training data for each workload, pick top-$m$ frequent $n$-gram sequences for each workload, and construct a set $\T{A}_n$ of the picked $n$-gram sequences collected from all known workloads.
Then, for each $\mat{S}_{i}(:,0)$, we construct a feature vector $\mat{c}_{i,n} \in \mathbb{R}^{|\T{A}_n|}$ whose entry is the number of occurrences for its corresponding element in $\T{A}_n$.
Fig.~\ref{fig:example_cmd_vectors} shows an example of generating a feature vector of command lines.

To capture sequential patterns of diverse lengths, we use several $n$-gram models for different $n$s.
In this paper, for each $\mat{S}_{i}(:,0)$, we use $7$, $11$ and $15$-gram models, generate three feature vectors $\mat{c}_{i,7}$, $\mat{c}_{i,11}$, and $\mat{c}_{i,15}$, and then construct a CMD feature vector $\mat{x}_{i, CMD} \in \mathbb{R}^{|\T{A}_7| + |\T{A}_{11}| + |\T{A}_{15}|}$ by concatenating $\mat{c}_{i,7}$, $\mat{c}_{i,11}$, and $\mat{c}_{i,15}$.

\textbf{ADDRESS Feature Vector for Address Related Fields.}
We next represent lines $\mat{S}_{i}$(:,1:4)$ \in \mathbb{R}^{100,000\times 4}$ of address-related fields (i.e., rank, bank group, bank, and address) of a subsequence as a feature vector $\mat{x}_{i,ADDRESS}$ of a small size.
\green{In address-related fields, it is important to capture the access pattern in memory.}
Therefore, we count how many times addresses are accessed in a subsequence. 
A naive approach is to compute access counts for each cell.
\green{However, this generates a large feature vector whose size is equal to $(\textit{\# of rank}) \times (\textit{\# of bank group}) \times (\textit{\# of bank}) \times (\textit{\# of address})$.}
To reduce the size of a feature vector for address-related fields, we
separately model bank-level access and cell-level access.
We also reduce the feature size for the cell-level access by defining memory regions and computing access counts for each region.

We first transform lines $\mat{S}_{i}$(:,1:3)$ \in \mathbb{R}^{100,000\times 3}$ of the rank, bank group, and bank fields into a feature vector $\mat{b}_i$.
Since there is a hierarchy for the three fields (see Fig.~\ref{fig:rank_address_hiearchy}) where the bank field is the lowest level, we generate a bank counting feature vector $\mat{b}_i$ by computing access counts in $\mat{S}_{i}$(:,1:3) for each distinguished bank.
For example, assume that there are $2$ ranks and each rank is a set of $4$ bank groups each of which has $4$ banks.
The total number of distinct banks is $32 = 2\times 4\times 4$.
Therefore, the size of a feature vector $\mat{b}_i$ is $32$, and an entry of the feature vector is the number of accesses for its corresponding bank.

\begin{figure*}[t]
	\centering
	\includegraphics[width=0.95\textwidth]{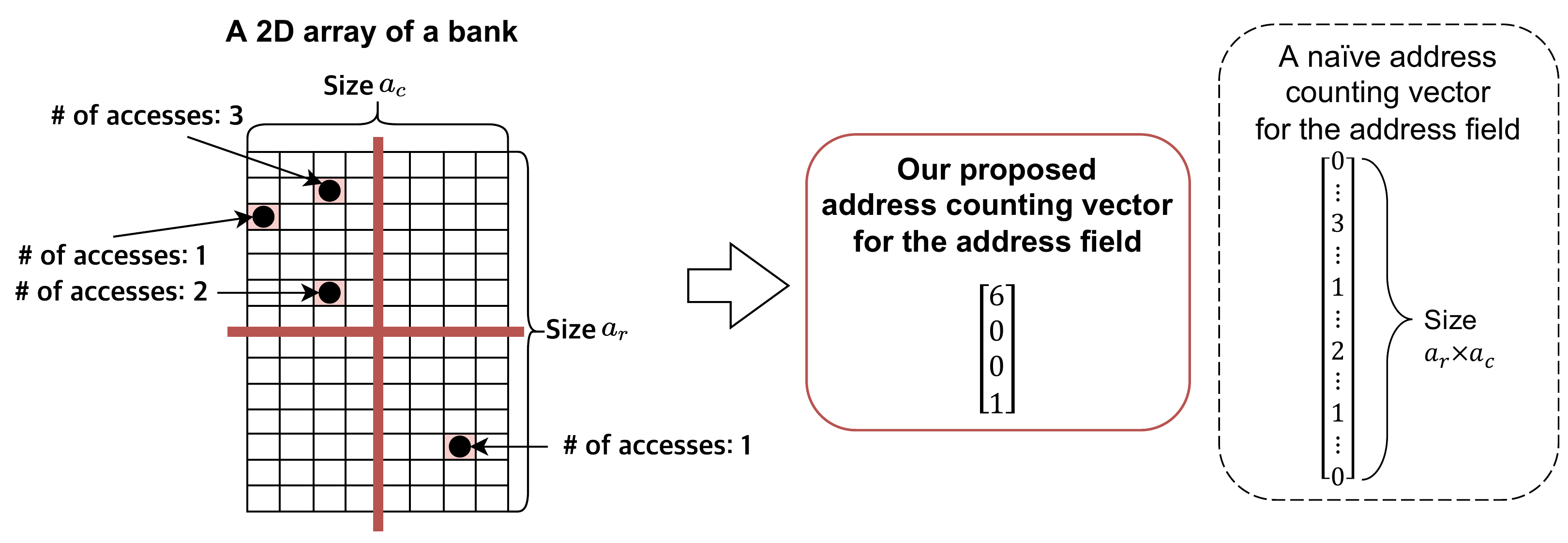}
	\vspace{-2mm}
	\caption{An example of constructing an address counting vector for the address field.
	We partition a 2D array into $4$ blocks by horizontal and vertical red lines, and then count the number of accesses in each block.
	}
	\label{fig:address_example}
\end{figure*}

We then transform address lines $\mat{S}_{i}(:,4) \in \mathbb{R}^{100,000\times 1}$ into a feature vector $\mat{d}_i$ of a small size.
A naive approach is to construct a feature vector by counting the number of accesses to each cell.
However, the size of this feature vector is equal to $(\textit{\# of distinct banks}) \times \textit{size of a bank}$ (e.g., $32 \times (2^{17} \times 2^{10})$).
Therefore, we partition each 2D array into smaller block regions and count the number of accesses to each block region.
\green{
A recent work~\cite{ZhangSNKP22} segments memory address to solve the problem of its high granularity.
Note that the difference between our address feature vector and the feature vector of the paper~\cite{ZhangSNKP22} is the information that an address feature vector contains.
Our address feature vector contains spatial information on the frequency of memory access in subsequence data, 
while the feature vector of the paper~\cite{ZhangSNKP22} contains sequential information on the time of the memory access.
}
For each distinct bank, we partition an address array of size $a_r \times a_c$ into $\frac{a_r}{g_r}\times \frac{a_c}{g_c}$ blocks of size $g_r \times g_c$, and generate a feature vector $\mat{d}_{i,b} \in \mathbb{R}^{\frac{a_r a_c}{g_r g_c}}$ by counting the number of accesses to blocks.
We provide an example in Fig.~\ref{fig:address_example}.
Then, we construct a feature vector $\mat{d}_{i} = \sum_{b=1} {\mat{d}_{i,b}}$ by summing up the feature vectors for all banks.
We transform $\mat{S}_{i}$(:,1:4) into an ADDRESS feature vector $\mat{x}_{i,ADDRESS} = \left(\mat{b}_{i} \| \mat{d}_{i}\right)$.

In summary, we transform all subsequences $\mat{S}_{i}$ into feature vectors $\mat{x}_{i}$ by concatenating $\mat{x}_{i, CMD}$ and  $\mat{x}_{i, ADDRESS}$, and exploit them for open-set recognition.

\subsection{Classification Model and Unknown Class Detector}
\textbf{Training Classification Model.}
Given a set $\{(\mat{S}_{1}, y_1), ..., (\mat{S}_{N}, y_N) \}$ of training samples, our goal is to learn a classification model for known workloads.
We construct a set $\{(\mat{x}_{1}, y_1), ..., (\mat{x}_{N}, y_N) \}$ by extracting feature vectors $\mat{x}_i$ from $\mat{S}_i$ for $i=1 .. N$, and then learn a classification model such as multi-layer perceptron.


\textbf{Unknown Class Detectors.}
After training a model, we build detectors that accurately identify whether test samples are the unknown class or not.
Given target size $R$ and a feature vector $\mat{x}' \in \mathbb{R}^{F}$ of a test sample where $F$ is the size of $\mat{x}'$, our strategy is to find a detector matrix $\mat{V} \in \mathbb{R}^{F \times R}$ based on training samples of known workloads, to measure a reconstruction error $\| \mat{x}' - \mat{V}\mat{V}^T\mat{x}'\|_2$.
\green{When $R$ is much smaller than $F$, the detector matrix maps a feature vector $\mat{x}'$ to a lower dimension and remaps it to a higher dimension.}
The intuition is that test samples of \emph{known} workloads generate low reconstruction errors while generating high reconstruction errors if test samples are from \emph{unknown} workloads.
A naive approach is to build one unknown workload detector for all known classes, but it fails to obtain a good decision boundary that distinguishes the unknown class from the known ones.
%
Our main ideas are to 1) build unknown workload detectors $\mat{V}_w$ for each class $w$, and 2) exploit feature vectors of training samples.
These ideas allow us to clearly distinguish test samples of known classes and the unknown class;
the difference in reconstruction errors between them becomes high.
Our approach is to find a detector matrix $\mat{V}_w$ for a known class $w$, which identifies whether a test sample belongs to the unknown class or the known class $w$ based on the reconstruction error $\| \mat{x}' - \mat{V}_{w}\mat{V}_{w}^T\mat{x}'\|_2$.
To obtain the matrix $\mat{V}_w$, we \green{utilize} Singular Value Decomposition (SVD) which is widely used in various applications including principal component analysis (PCA)~\cite{jolliffe2002principal,wall2003singular}, data clustering~\cite{simek2004using,osinski2004lingo}, tensor analysis~\cite{JangK20}, and time range analysis~\cite{JangCJK18}.
With SVD, $\mat{V}_w$ minimizes the reconstruction error when a test sample $x'$ belongs to the known class $w$.
In contrast to the test samples of the known class $w$, a reconstruction error for a test sample of the unknown class is high since the characteristics of the known class $w$ are different from those of the unknown class.
\green{For each class $w$, we construct a matrix $\mat{X}_{w} \in \mathbb{R}^{N_{w} \times F}$ where each row corresponds to a feature vector of a training sample belonging to the class $w$.}
Note that $N_{w}$ is the number of training samples of the class $w$.
Then, we perform SVD for the matrix $\mat{X}_{w} = \mat{U}_{w}\mat{\Sigma}_{w}\mat{V}_{w}^T$ and obtain the matrix $\mat{V}_{w} \in \mathbb{R}^{F \times R_w}$ of the right singular vectors where $R_w$ is target rank for the class $w$, to exploit it as a detector.
With $\mat{V}_{w}$, we clearly identify whether a subsequence belongs to a known class $w$ or the unknown class by measuring a reconstruction error $\| \mat{x}' - \mat{V}_{w}\mat{V}_{w}^T\mat{x}'\|_2$.
$\mat{x}'$ is a feature vector of a test sample.

\begin{algorithm} [t]
	\caption{\method: Open-set Recognition for Workload Sequence}
	\label{alg:method}
	\begin{algorithmic} [1]
		\renewcommand{\algorithmicrequire}{\textbf{Input:}}
		\renewcommand{\algorithmicensure}{\textbf{Output:}}
		    \REQUIRE A test sample $\mat{S}'$, a trained classification model, a threshold \green{hyperparameter} $\alpha$, \newline unknown class detector matrices $\mat{V}_{w} \in \mathbb{R}^{F \times R_w}$, $\mu_w$, and $\sigma_w$ for $w=1,...,W$ where $W$ is the number of known classes  \\
		    \ENSURE One of the known class labels or the unknown class label for $\mat{S}'$\\
		\STATE {\textbf{Feature Extraction.} Transform $\mat{S}'$ into a feature vector $\mat{x}' \in \mathbb{R}^{F}$ using our feature \newline extraction approach in Section~\ref{subsec:cons_vectors}.}
		\STATE {\textbf{Known Class Prediction.} Predict a class label $\hat{w}$ using the trained classification model.}
		\STATE {\textbf{Unknown Class Detection.} Identify whether to satisfy $\| \mat{x}' - \mat{V}_{\hat{w}}\mat{V}_{\hat{w}}^T\mat{x}'\|_2 < \mu_{\hat{w}} + \alpha * \sigma_{\hat{w}}$ \newline for the $\hat{w}$th class.
		If the above inequality condition is satisfied, identify it as the predicted \newline class label $\hat{w}$.
		Otherwise, identify it as the unknown class label.}
	\end{algorithmic}
\end{algorithm}

\subsection{Open-set Recognition for Workload Sequence}
We describe how to identify a test sample $\mat{S}'$ as the unknown class using a trained classification model and SVD-based detectors (Algorithm~\ref{alg:method}).
Given a feature vector $\mat{x}'$ of the test sample, we first predict the class label $\hat{w}$ using the trained classification model where $\mat{x}'$ is extracted from $\mat{S}'$.
After that, we recognize whether the test sample belongs to the predicted label or not by computing a reconstruction error with $\mat{V}_{\hat{w}}$.
We recognize it as the predicted class label $\hat{w}$ only when the following inequality condition is satisfied.
\begin{align}
	\| \mat{x}' - \mat{V}_{\hat{w}}\mat{V}_{\hat{w}}^T\mat{x}'\|_2 < \epsilon_{\hat{w}}
\label{eq:recon}
\end{align}
where $\mat{V}_{\hat{w}}\mat{V}_{\hat{w}}^T\mat{x}'$ is the reconstructed vector.
\green{When Eq.~\eqref{eq:recon} is not satisfied, we determine that the test sample belongs to the unknown class.}
We set a threshold $\epsilon_{\hat{w}}$ to $\mu_{\hat{w}} + \alpha * \sigma_{\hat{w}}$ where $\mu_{\hat{w}}$ and $\sigma_{\hat{w}}$ are the mean and the standard deviation of reconstruction errors for feature vectors of training samples of the class $\hat{w}$, respectively.
$\alpha$ provides a trade-off between known class classification accuracy and unknown class detection accuracy.
A high $\alpha$ increases known class classification accuracy, but decreases unknown class detection accuracy.
This is because false negatives increase while false positives decrease for the unknown class.
A low $\alpha$ does the opposite.


\section{Experiments}
\label{sec:experiment}


In this section, we experimentally evaluate the performance of \method.
We aim to answer the following questions:

\begin{itemize}[noitemsep,topsep=0pt]
	\item[Q1.] \textbf{Performance (Section 4.2).} How accurately does \method classify subsequences from known workloads and detect subsequences from unknown workloads?
	\item[Q2.] \textbf{Feature Effectiveness (Section 4.3).} How successfully do our feature vectors improve the classification accuracy?
	\item[Q3.] \textbf{Effectiveness of Per-Class Detector (Section 4.4).}
How accurately do per-class detectors identify the unknown class compared to the naive detector?
\end{itemize}

\subsection{Experimental Setting}
\label{subsec:experimental_setting}
\green{We construct our model using the Pytorch framework.
All the models are trained and tested on a machine with a GeForce GTX 1080 Ti GPU.}

\textbf{Datasets.}
We use two real-world workload sequence datasets summarized in Table~\ref{tab:data}.
There are $44$ and $34$ workloads for \fdata\footnote{Private to a company.\label{foot:samsung_data}} and Memtest86-seq\footnote{\url{https://github.com/snudatalab/Acorn}\label{foot:memtest}}, respectively.
\green{We publicize Memtest86-seq dataset which is generated from an open-source software program for DRAM test.
Memtest86 uses two different algorithms to find out memory errors which are often caused by interaction between memory cells.
For a workload, we collect signals using an equipment capturing DRAM signal, and transform the signals into a sequence with $5$ heterogeneous fields in Definition 1.}
The lengths of each workload sequence are different.

\textbf{DRAM Specification.}
Both \fdata and Memtest86-seq datasets are generated on a server with Samsung 32GB 2Rx4 2666Mhz DRAM chip and Intel Gold6248 CPU.
The DRAM chip has 2 ranks, and there are $4$ bank groups in a rank and $4$ banks in a bank group so that there are $2\times 4 \times 4 = 32$ distinct banks.
Each bank has $2^{17}$ rows and $2^{10}$ columns with a cell size of $8$ Bytes which gives total of $2^{17} \times 2^{10} \times 8$ Bytes = $1$GB.

\setlength{\tabcolsep}{6pt}
\begin{table*}[t]
\caption{Datasets. \textit{known} and \textit{unknown} are the short term for known workloads and unknown workloads, respectively. The number of known workloads is equal to the number of known classes. All unknown workloads correspond to one unknown class.}
\centering
\vspace{-2mm}
\label{tab:data}
\resizebox{ \columnwidth}{!}{%
\begin{tabular}{lrrrrr}
\toprule
 & \textbf{\# of \textit{known}}  & \textbf{\# of \textit{unknown}}  & \textbf{\# of} & \textbf{\# of test} & \textbf{\# of test} \\
\textbf{Dataset} & \textbf{workloads} & \textbf{workloads} & \textbf{train} & \textbf{of \textit{known}} & \textbf{ of \textit{unknown}} \\
\midrule
SEC-seq\footnoteref{foot:samsung_data} & $40$ & $4$ & $586,885$ & $293,444$ & $93,491$\\
Memtest86-seq\footnoteref{foot:memtest} & $31$ & $3$  & $433,334$ & $216,696$ & $77,018$ \\
\bottomrule
\end{tabular}}
\end{table*}

\textbf{Competitors for open-set recognition.}
We compare \method with the following $9$ competitors for open-set recognition:
	\begin{itemize}[noitemsep,topsep=0pt]
		\item \textbf{Naive Rejection~\cite{HendrycksG17}} identifies a test sample as the unknown class when the maximum softmax score of a trained model is below a threshold.
		\item \textbf{OpenMax~\cite{BendaleB16}} adds a new class called \textit{unknown} and applies softmax with a threshold.
		\item \textbf{ii-loss~\cite{HassenC20}} forces a network to maximize the distances between known classes and minimize the distance between an instance and the center of its class.
		\item \textbf{DOC~\cite{ShuXL17}} changes \green{the softmax layer} to 1-vs.-rest layer with sigmoid function.
		\item \textbf{ODIN~\cite{LiangLS18}} adds a small perturbation to the input and divides softmax values by temperature parameter $T$.
		\item \textbf{Mahalanobis-based detector~\cite{LeeLLS18}} computes \green{the confidence score} using Mahalanobis distance instead of Euclidean distance.
		\item \textbf{Deep-MCDD~\cite{LeeYY20}} obtains spherical decision boundary for each given class and computes the distances of samples from each class.
		\item \textbf{Energy-based detector~\cite{LiuWOL20}} uses energy scores to differentiate the out-of-distribution data from the in-distribution data.
		\item \textbf{ReAct~\cite{Sun21}} applies rectified activation on the penultimate layer of a network and calculates the confidence score.
	\end{itemize}

We use $2$-layer MLP as a classification model, and all the methods are combined with the MLP.
\green{
In addition, \method and competitors use the same input feature vectors.
This is because it is impracticable for using a subsequence of size $100,000\times 5$ not processed by our feature extraction method to learn competitors.
Our feature extraction method enables us to learn models effectively on our open-set recognition task.
}

\textbf{Hyperparameter Settings.}
\green{
We use the following \green{hyperparameters} in experiments:
\begin{itemize}
 \item \textbf{\green{Hyperparameters} for $n$-gram CMD vectors.}
\begin{itemize}
 \item We construct three $n$-gram CMD vectors for each subsequence: $n = 7,11,$ and $15$.
 \item In order to generate a set $\T{A}_n$, we select top-$m$ $n$-gram CMD vectors for each $n$ with $m = 25$.
 \item For \fdata, $| \T{A}_7 |$, $| \T{A}_{11} |$, and $| \T{A}_{15} |$ are $154$, $236$, and $289$, respectively.
For Memtest86-seq, $| \T{A}_7 |$, $| \T{A}_{11} |$, and $| \T{A}_{15} |$ are $132$, $196$, and $215$, respectively.
\end{itemize}
 \item \textbf{\green{Hyperparameters} for address counting vectors.}
\begin{itemize}
 \item We split an address array of size $2^{17} \times 2^{10}$ into $8\times 128$ blocks of size $2^{14} \times 8$ where $g_r$ and $g_c$ are equal to $2^{14}$ and $8$, respectively.
\end{itemize}
 \item \textbf{\green{Hyperparameters} for our open-set recognition model.}
\begin{itemize}
\item For multi-layer perceptron (MLP), we use Adam optimizer with a learning rate of $0.0001$ and fix the batch size to $128$.
\item The total number of epochs is set to $5$.
\item We set $\alpha$ as one of $\{1,1.5,2,2.5,3\}$ to compute the threshold of \method, and select thresholds of the competitors based on their papers.
\item The rank $R_w$ is set to the minimum value $r$ that satisfies $\sum_{k=1}^{r}{\lambda^2_{k}}> 0.999 \times \sum_{k=1}^{\min{(N_w, F)}}{\lambda^2_{k}}$ for a class $w$ where $\lambda_{k}$ is the $k$th singular value of the SVD result for $\mat{X}_w$.
\end{itemize}
\end{itemize}
}


\textbf{Evaluation Metrics.}
Accuracy $(\%)$ for \green{known classes} is equal to $(\bar{N}'_{known}/N'_{known}) \times 100$ where $N'_{known}$ is the number of test samples of known classes and $\bar{N}'_{known}$ is the number of test samples correctly classified as true known classes.
The metrics of precision $(\%)$ and recall $(\%)$ for the unknown class are also used for unknown class detection.
\green{We compute f1-score $(\%)$ to show the trade-off between the unknown class recall and unknown class precision.}

\begin{figure*}[!t]
	\centering
	\vspace{-3mm}
	 \subfloat{\includegraphics[width=1\textwidth]{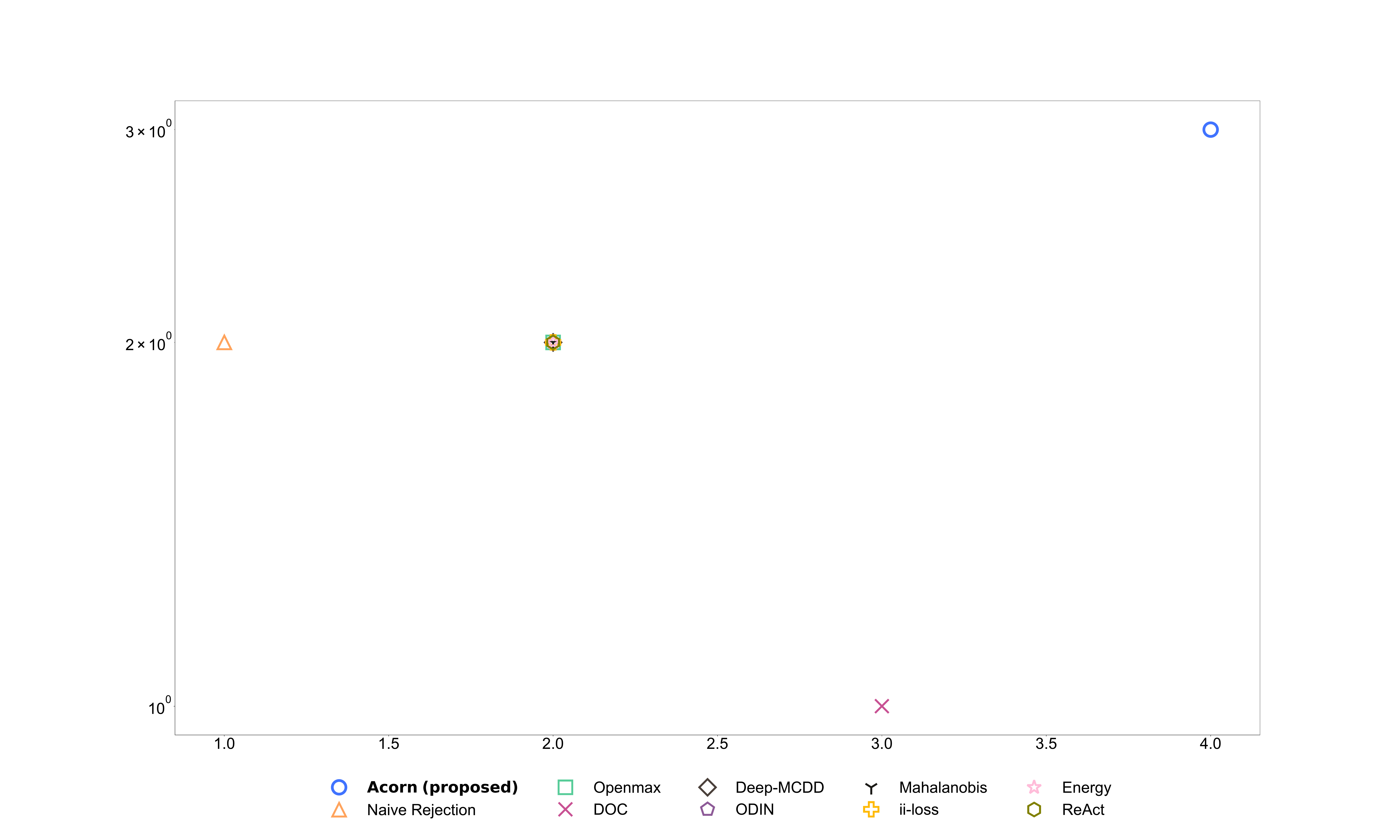}} \\
	\vspace{-3mm}	
	 \setcounter{subfigure}{0}
	 	\centering
	 \subfloat[\fdata]{\includegraphics[width=0.225\textwidth]{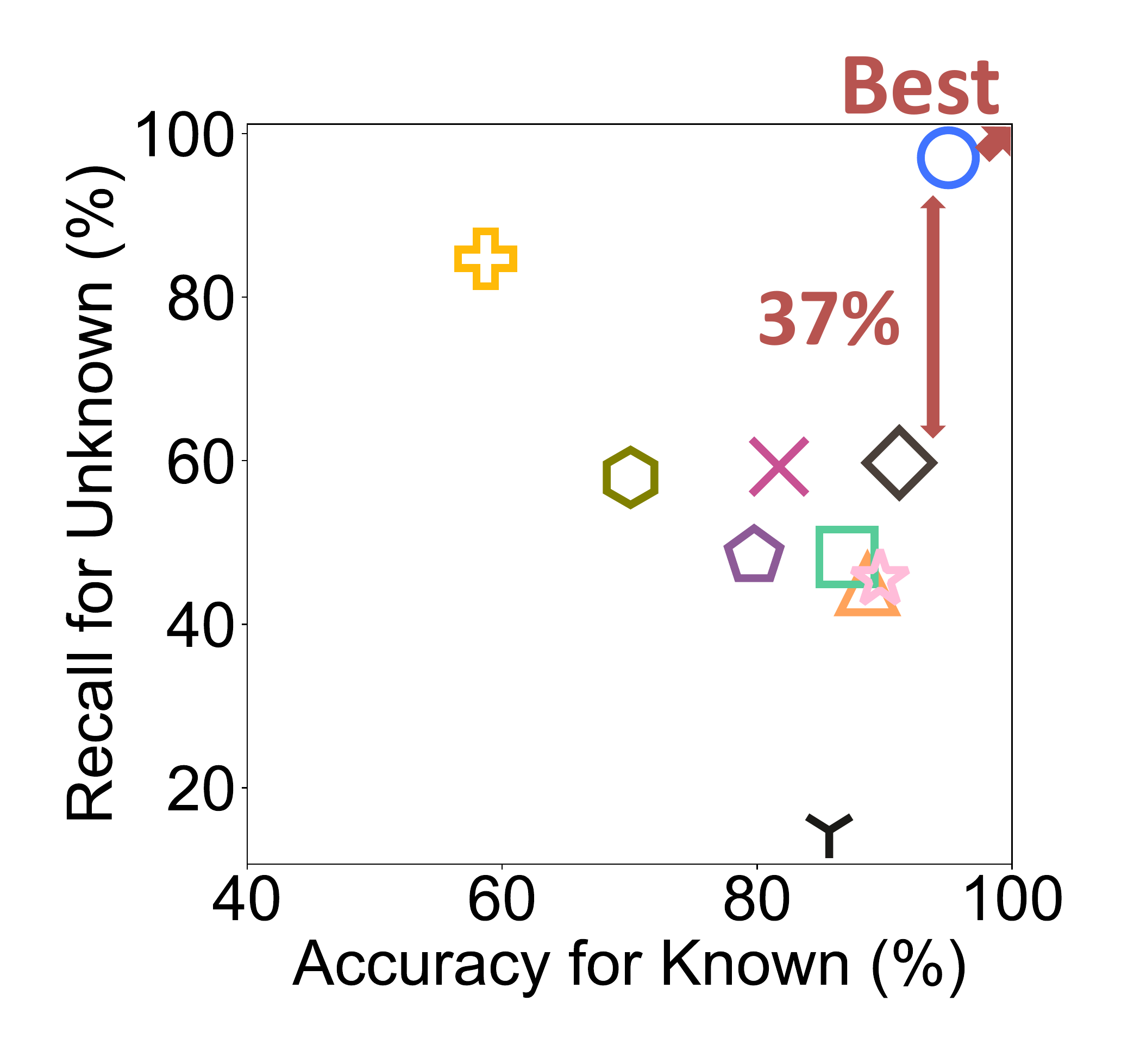}\label{fig:trade_off1}}
	 \subfloat[SEC-seq]{\includegraphics[width=0.225\textwidth]{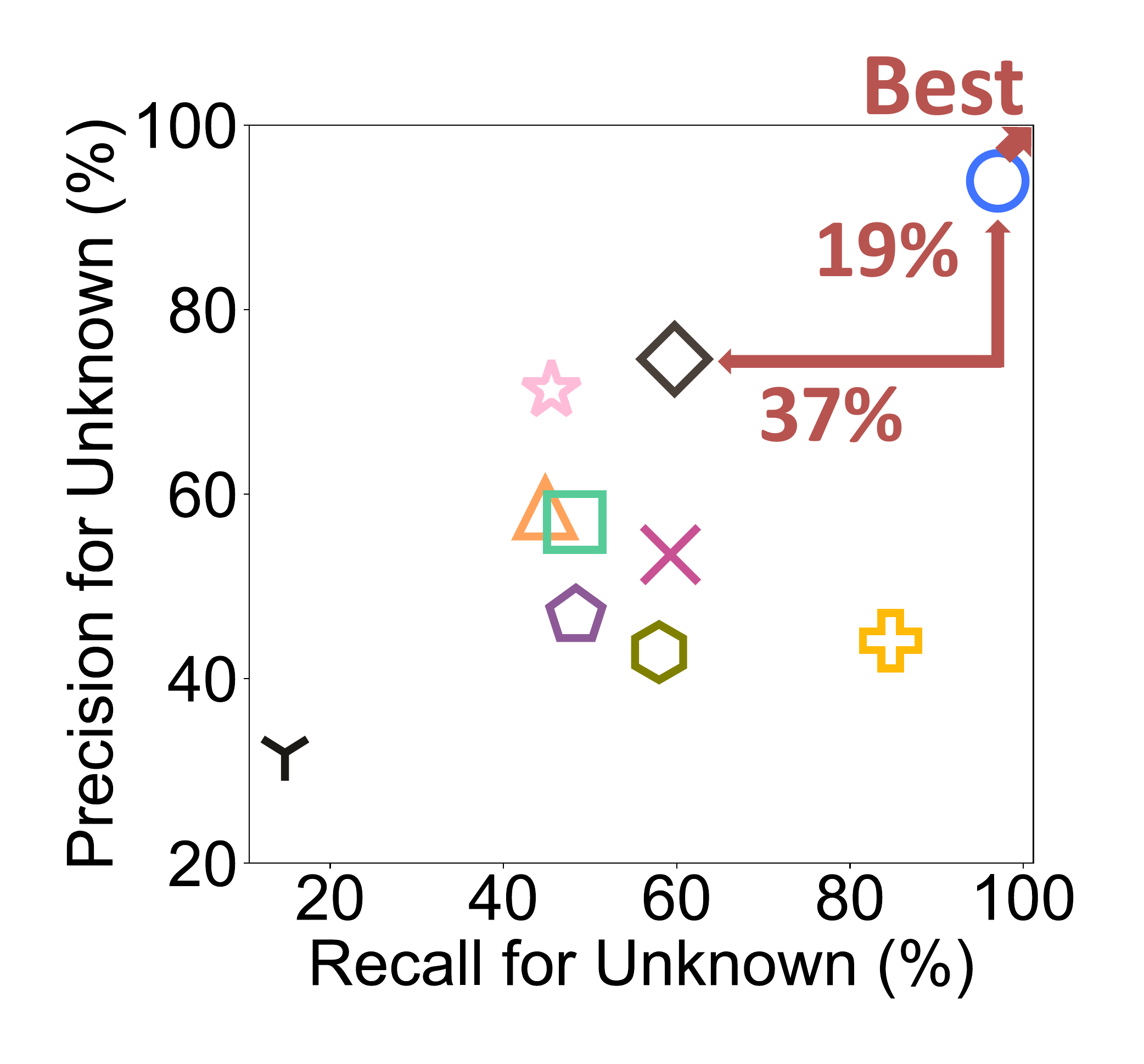}\label{fig:trade_off2}}		
	 \subfloat[Memtest86-seq]{\includegraphics[width=0.23\textwidth]{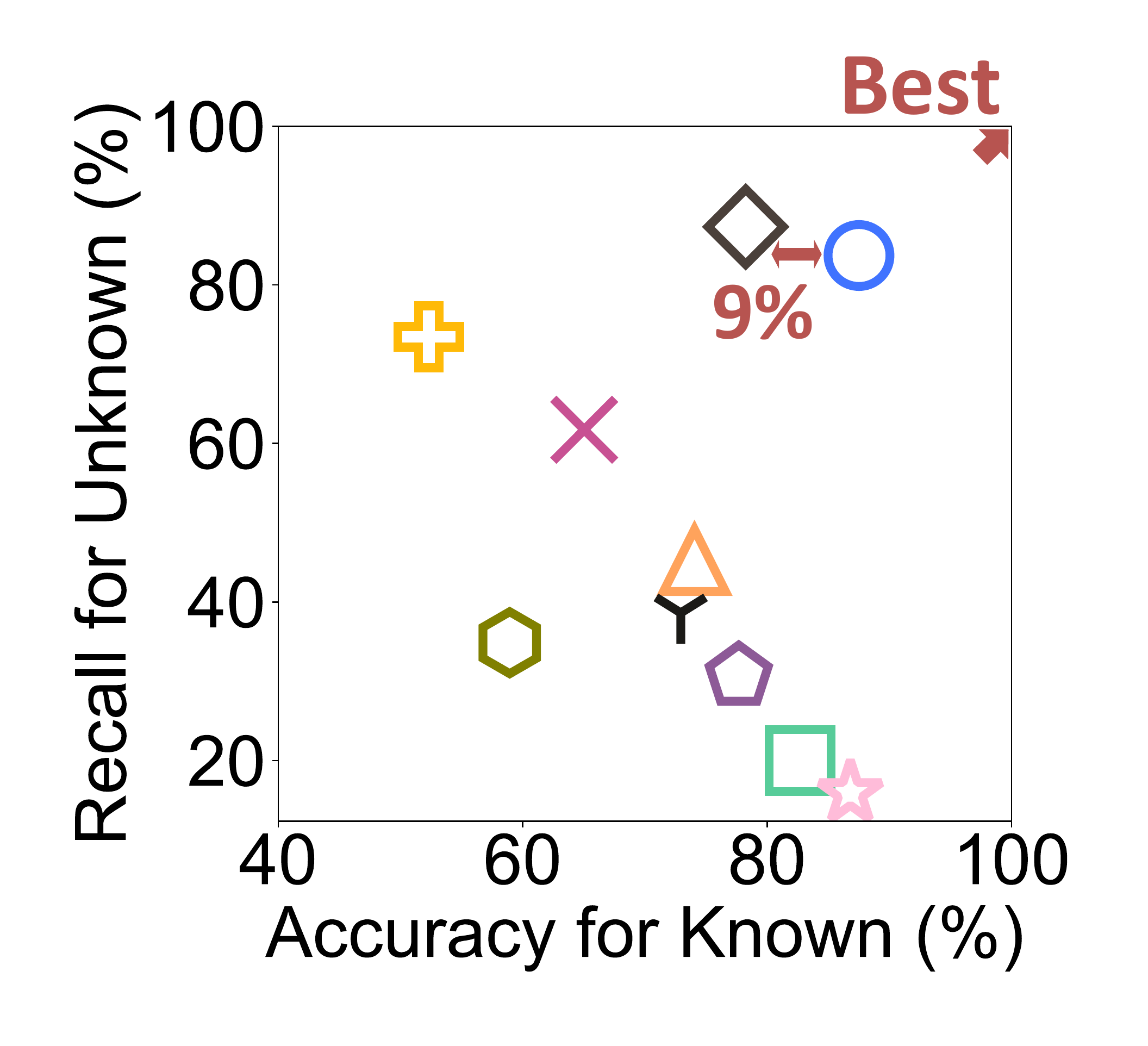}\label{fig:trade_off3}}
	 \subfloat[Memtest86-seq]{\includegraphics[width=0.23\textwidth]{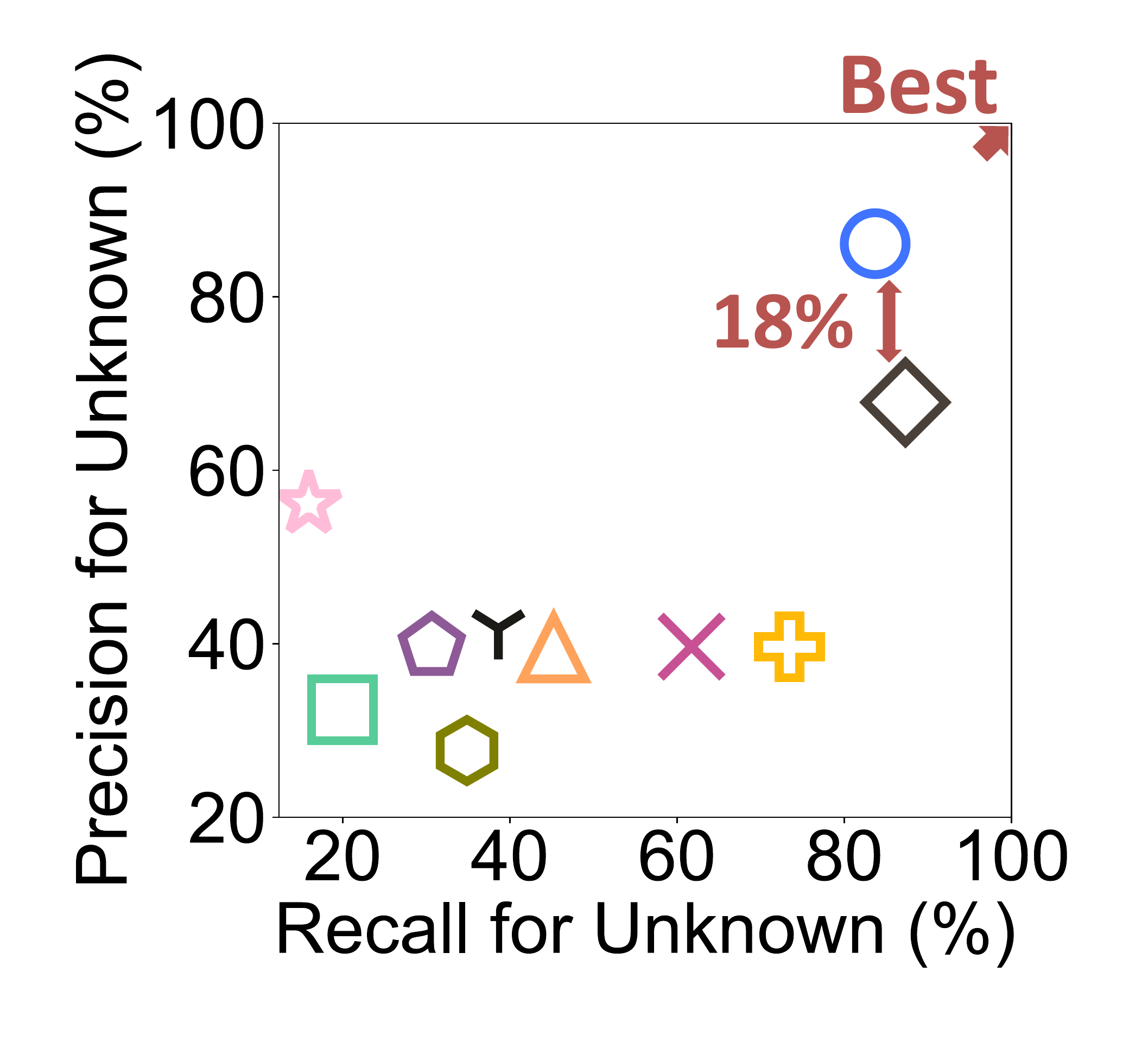}\label{fig:trade_off4}}
	 	 \\
	 	\vspace{-1mm}
	\caption{
	Performance for open-set recognition. Note that the best results of each method are reported. \method provides the best trade-off between accuracy for known class classification and precision \& recall for unknown class detection.
	}
	\label{fig:performance}
\end{figure*}

\subsection{Performance}
\label{subsec:performance}
\green{In this section, we show the performance of \method in terms of known class classification accuracy, unknown class recall, unknown class precision, and inference time.}
\vspace{-2mm}

\green{
\subsubsection{Accuracy}
We compare the performance of \method with competitors on the open-set recognition task.
We observe known class classification accuracy, unknown class recall, and unknown class precision.
We report F1-score for unknown class detection to show the trade-off between the precision and recall.}
Fig.~\ref{fig:performance} visualizes the best trade-off between the evaluation metrics of the proposed method and the competitive methods.
For both datasets, \method provides the best performance compared to various open-set recognition methods.
In Fig.~\ref{fig:trade_off1} and ~\ref{fig:trade_off2}, \method achieves $37\%$ points higher recall and $19\%$ points higher precision for the unknown class than the second-best competitor Deep-MCDD.
In Fig.~\ref{fig:trade_off3} and ~\ref{fig:trade_off4}, \method gives $9\%$ points higher known class classification accuracy and $18\%$ points higher precision for the unknown class than the second-best method while having a similar recall.
We also report the overall results of the proposed method and the competitors for both datasets in Table~\ref{tab:performance_sec}.
\green{As for ii-loss method, the best known class accuracy is lower than $95\%$ points and $90\%$ points for SEC-seq and Memtest86-seq datasets, respectively; thus, we do not report the performance for these cases.
\green{\method shows the highest unknown class f1-score with fixed known class accuracy for all cases.}
The performance gap occurs since \method constructs an unknown class detector by exploiting feature vectors, while competitors use hidden vectors generated from known classification models.
The hidden vectors do not contain enough information to detect the unknown class since known classification models concentrate only on extracting information that classifies an input into known ones.
}
\vspace{-2mm}

\begin{table}[!t]
\caption{\green{Performance for the open-set recognition. acc., rec., prec., and f1 denote known class accuracy, unknown class recall, unknown class precision, and unknown class f1-score, respectively. Bold and underlined text represent the best and the second-best performance, respectively.}}
\vspace{-2mm}
\centering
\label{tab:performance_sec}
\resizebox{\columnwidth}{!}{%
\begin{tabular}{l|cccccc|cccccc}
	\toprule
	\multirow{3}{*}{\textbf{Method}}
	& \multicolumn{6}{c|}{\textbf{SEC-seq}} & \multicolumn{6}{c}{\textbf{Memtest86-seq}}
	\\
	\cline{2-13}
	& \multicolumn{3}{c|}{acc. = 95} & \multicolumn{3}{c|}{acc. = 85}
	& \multicolumn{3}{c|}{acc. = 90} & \multicolumn{3}{c}{acc. = 80}
	\\
	\cline{2-13}
	& rec. & \multicolumn{1}{c}{prec.} & \multicolumn{1}{c|}{f1}
	& rec. & \multicolumn{1}{c}{prec.} & \multicolumn{1}{c|}{f1}
	& rec. & \multicolumn{1}{c}{prec.} & \multicolumn{1}{c|}{f1}
	& rec. & \multicolumn{1}{c}{prec.} & \multicolumn{1}{c}{f1}
	\\
	\midrule
	Naive Rejection & \multicolumn{1}{c}{19.66} & 64.53 & 30.14
	& \multicolumn{1}{|c}{48.92} & 54.41 & 51.52
	& \multicolumn{1}{c}{\underline{16.81}} & 60.35 & 26.30
	& \multicolumn{1}{|c}{44.13} & 46.42 & 45.25 \\
	OpenMax & \multicolumn{1}{c}{\underline{29.02}} & 82.73 & \underline{42.97}
	& \multicolumn{1}{|c}{48.97} & 52.90 & 50.86
	& \multicolumn{1}{c}{12.35} & 58.07 & 20.37
	& \multicolumn{1}{|c}{25.70} & 32.92 & 28.87 \\
	ii-loss & \multicolumn{1}{c}{---} & --- & ---
	& \multicolumn{1}{|c}{14.19} & 36.97 & 20.51
	& \multicolumn{1}{c}{---} & --- & ---
	& \multicolumn{1}{|c}{18.78} & 50.72 & 27.41 \\
	DOC & \multicolumn{1}{c}{23.10} & 84.19 & 36.25
	& \multicolumn{1}{|c}{58.18} & 58.24 & 58.21
	& \multicolumn{1}{c}{3.17} & 86.19 & 6.12
	& \multicolumn{1}{|c}{24.22} & 35.35 & 28.75 \\
	ODIN & \multicolumn{1}{c}{16.56} & 75.16 & 27.14
	& \multicolumn{1}{|c}{50.17} & 55.49 & 52.70
	& \multicolumn{1}{c}{8.06} & 70.29 & 14.46
	& \multicolumn{1}{|c}{12.04} & 24.75 & 16.20 \\
	Mahalanobis & \multicolumn{1}{c}{5.74} & 28.72 & 9.57
	& \multicolumn{1}{|c}{11.62} & 19.72 & 14.62
	& \multicolumn{1}{c}{3.14} & 27.71 & 5.64
	& \multicolumn{1}{|c}{13.41} & 19.91 & 16.03 \\
	Deep-MCDD & \multicolumn{1}{c}{19.28} & \textbf{94.96} & 32.05
	& \multicolumn{1}{|c}{\underline{61.13}} & \textbf{73.26} & \underline{66.65}
	& \multicolumn{1}{c}{16.39} & \underline{90.09} & \underline{27.74}
	& \multicolumn{1}{|c}{\underline{83.23}} & \underline{70.49} & \underline{76.33} \\
	Energy & \multicolumn{1}{c}{23.21} & 78.54 & 35.83
	& \multicolumn{1}{|c}{59.44} & 71.48  & 64.91
	& \multicolumn{1}{c}{9.01} & 51.63 & 15.34
	& \multicolumn{1}{|c}{19.08} & 43.59 & 26.54 \\
	ReAct & \multicolumn{1}{c}{16.94} & 75.98 & 27.70
	& \multicolumn{1}{|c}{17.59} & 38.99 & 24.24
	& \multicolumn{1}{c}{7.27} & 61.83 & 13.01
	& \multicolumn{1}{|c}{7.36} & 20.53 & 10.84
	\\
	\midrule
	\textbf{\method} & \multicolumn{1}{c}{\textbf{97.05}} & \underline{93.96} & \textbf{95.48}
	& \multicolumn{1}{|c}{\textbf{99.88}} & \underline{73.15} & \textbf{84.45}
	& \multicolumn{1}{c}{\textbf{60.33}} & \textbf{91.71} & \textbf{72.78}
	& \multicolumn{1}{|c}{\textbf{95.11}} & \textbf{72.41} & \textbf{82.22}
	\\
	\bottomrule
\end{tabular}}
\end{table}

\green{
\subsubsection{Inference time}
We compare the inference time of the proposed method with competitors in Table~\ref{tab:inference_time}.
We measure the CPU running time of the inference procedure.
\method takes the second shortest inference times which are $46.47$ and $24.03$ seconds slower than DOC for SEC-seq and Memtest86-seq datasets, respectively.
\green{However, the proposed method shows better performance than DOC for all cases.
\method achieves $59.23\%$ points higher unknown class f1-score than DOC when the known class accuracy is fixed as $95\%$ points.}
For Memtest86-seq, \method achieves $66.66\%$ points higher f1-score than DOC, while having the same accuracy.
\method is the only method to achieve high accuracy and fast inference simultaneously on the workload open-set recognition task.
}
\begin{table}[!t]
\caption{\green{Performance for the inference time. Naive, Mahal., and MCDD are abbreviations for Naive Rejection, Mahalanobis, and Deep-MCDD methods, respectively. We report the fastest time as bold and the second fastest time as underline.}}
\centering
\label{tab:inference_time}
\resizebox{\columnwidth}{!}{%
\begin{tabular}{l|cccccccccc}
	\toprule
	\textbf{Dataset} & Naive & OpenMax & ii-loss & DOC & ODIN & Mahal. & MCDD & Energy & ReAct & \method
	\\
	\midrule
	\textbf{SEC-seq} & 250.60 & 548.35 & 225.23 & \textbf{29.68} & 80.65 & 2852.02 & 635.80 & 773.28 & 1731.35 & \underline{76.15}
	\\
	\textbf{Memtest86-seq} & 155.25 & 372.16 & 138.18 & \textbf{31.94} & 77.98 & 1109.93 & 378.44 & 539.98 & 1608.12 & \underline{55.97}
	\\
	\bottomrule
\end{tabular}}
\end{table}

\subsection{Feature Effectiveness}
We evaluate models for 6 different feature vectors: CMD (only $7$-gram), CMD (only $11$-gram), CMD (only $15$-gram), CMD (the concatenation of $7,11$, and $15$-grams CMD vectors), ADDRESS, and CMD + ADDRESS (the concatenation of the $7,11$, and $15$-grams CMD vectors, and the ADDRESS vector).
We evaluate classification accuracy for known classes.
Table~\ref{tab:feature_effectiveness} shows the results for both datasets.
Our CMD + ADDRESS feature vectors make the model (MLP) achieve the highest classification accuracy.
For both datasets, ADDRESS feature vectors are more effective than CMD vectors,
since ADDRESS vectors give more detailed information on the target address than CMD vectors do.
In addition, using several $n$-gram models provides higher accuracy than using one $n$-gram model.

\setlength{\tabcolsep}{4pt}
\begin{table*}[!t]
\caption{Feature effectiveness. Our feature vectors (CMD + ADDRESS) make the classifier model (MLP) achieve the highest accuracy for known classes.}
\centering
\vspace{-2mm}
\label{tab:feature_effectiveness}
\resizebox{\columnwidth}{!}{%
\begin{tabular}{lrrrrrr}
\toprule
                 & CMD                & CMD                 & CMD               &      &    & \textbf{CMD +} \\
\textbf{Dataset} & (only 7-gram)  & (only 11-gram)  & (only 15-gram) & CMD & ADDRESS & \textbf{ADDRESS} \\
\midrule
\fdata & $82.20 \%$ & $81.92\%$ & $80.13\%$ & $83.56\%$ & $90.88\%$ & $\mathbf{97.02\%}$\\
Memtest86-seq & $65.93\%$ & $62.68\%$  & $57.54\%$ & $67.42\%$ & $76.29\%$ & $\mathbf{92.08\%}$ \\
\bottomrule
\end{tabular}}
\end{table*}

\begin{table}[!t]
\caption{Comparison between the naive method and \method for unknown workload detection.
$\alpha$ is a \green{hyperparameter} for thresholds.
Note that acc., rec., and prec. denote known class accuracy, unknown class recall, and unknown class precision, respectively.
See Section~\ref{subsec:effective_perclass} for details.
}
\vspace{-2mm}
\centering
\resizebox{\columnwidth}{!}{%
\begin{tabular}{l|crrcrrcrrcrrcrr}
\toprule
\multirow{2}{*}{\textbf{SEC-seq}} & \multicolumn{3}{c|}{$\alpha$ = 1}                                                              & \multicolumn{3}{c|}{$\alpha$ = 1.5}                                                            & \multicolumn{3}{c|}{$\alpha$ = 2}                                                              & \multicolumn{3}{c|}{$\alpha$ = 2.5}                                                            & \multicolumn{3}{c}{$\alpha$ = 3}                                                              \\
\cline{2-16}
                         & acc.                  & \multicolumn{1}{c}{rec.} & \multicolumn{1}{c|}{prec.} & acc.                  & \multicolumn{1}{c}{rec.} & \multicolumn{1}{c|}{prec.} & acc.                  & \multicolumn{1}{c}{rec.} & \multicolumn{1}{c|}{prec.} & acc.                  & \multicolumn{1}{c}{rec.} & \multicolumn{1}{c|}{prec.} & acc.                  & \multicolumn{1}{c}{rec.} & \multicolumn{1}{c}{prec.} \\
\midrule
Naïve                    & \multicolumn{1}{r}{81.80} & 47.27                      & 51.23                         & \multicolumn{1}{|r}{87.23} & 29.44                      & 50.59                         & \multicolumn{1}{|r}{91.44} & 19.41                       & 54.34                         & \multicolumn{1}{|r}{94.20} & 12.10                       & 59.65                         & \multicolumn{1}{|r}{95.74} & 7.95                       & 68.32                         \\
\textbf{\method}                    & \multicolumn{1}{r}{86.19} & 99.86                      & 75.35                         & \multicolumn{1}{|r}{90.79} & 99.5                       & 83.97                         & \multicolumn{1}{|r}{93.04}    & 98.83                      & 89.00                         & \multicolumn{1}{|r}{94.27} & 98.04                      & 92.05                         & \multicolumn{1}{|r}{$\mathbf{95.00}$}    & $\mathbf{97.05}$                      & $\mathbf{93.96}$                   \\
\bottomrule
\end{tabular}}

\resizebox{\columnwidth}{!}{%
\begin{tabular}{l|crrcrrcrrcrrcrr}
\toprule
\multirow{2}{*}{\textbf{Memtest86-seq}} & \multicolumn{3}{c|}{$\alpha$ = 1}                                                              & \multicolumn{3}{c|}{$\alpha$ = 1.5}                                                            & \multicolumn{3}{c|}{$\alpha$ = 2}                                                              & \multicolumn{3}{c|}{$\alpha$ = 2.5}                                                            & \multicolumn{3}{c}{$\alpha$ = 3}                                                              \\
\cline{2-16}
                         & acc.                  & \multicolumn{1}{c}{rec.} & \multicolumn{1}{c|}{prec.} & acc.                  & \multicolumn{1}{c}{rec.} & \multicolumn{1}{c|}{prec.} & acc.                  & \multicolumn{1}{c}{rec.} & \multicolumn{1}{c|}{prec.} & acc.                  & \multicolumn{1}{c}{rec.} & \multicolumn{1}{c|}{prec.} & acc.                  & \multicolumn{1}{c}{rec.} & \multicolumn{1}{c}{prec.} \\
\midrule
Naïve                    & \multicolumn{1}{r}{76.99} & 78.78                      & 64.21                         & \multicolumn{1}{|r}{83.68} & 70.76                      & 74.62                         & \multicolumn{1}{|r}{88.62} & 60.37                       & 86.60                         & \multicolumn{1}{|r}{90.64} & 49.14                       & 93.67                         & \multicolumn{1}{|r}{91.25} & 39.85                       & 96.18                         \\
\textbf{\method}                    & \multicolumn{1}{r}{83.12} & 92.25                      & 77.67                         & \multicolumn{1}{|r}{$\mathbf{87.52}$} & $\mathbf{83.70}$                       & $\mathbf{86.12}$                         & \multicolumn{1}{|r}{89.36}    & 72.57                      & 90.04                         & \multicolumn{1}{|r}{90.22} & 60.33                      & 91.71                         & \multicolumn{1}{|r}{90.67}    & 48.23                      & 92.21                   \\
\bottomrule
\end{tabular}}
\label{tab:detector}
\end{table}

\subsection{Effectiveness of Per Class Detector}
\label{subsec:effective_perclass}
We compare \method and the naive SVD detector that does not consider classes.
\begin{itemize}[noitemsep,topsep=0pt]
	\item \textbf{Naive}: we obtain one $\mat{V}$ by computing SVD \green{for all training samples,} and then recognize a test sample using $\mat{V}$ by computing a reconstruction error.
\end{itemize}
Table~\ref{tab:detector} shows that \method achieves much higher performance than the naive SVD detector.
For \fdata, \method outperforms the naive detector \green{for all $\alpha$.}
At $\alpha = 3$, \method gives $89\%$ points higher recall
and $25\%$ points higher precision
than the naive one while having a comparable known class accuracy.
For Memtest86-seq, \method and the naive detector have comparable known class accuracies and precision, but \method gives at least $8.3\%$ points higher recall than the naive one.
Since feature vectors of classes have different patterns, the single $\mat{V}$ of the naive method fails to have the capacity to distinguish known classes and the unknown class.

\section{Conclusion}
\label{sec:conclusion}

In this paper, we propose \method, an accurate open-set recognition method for workload sequences.
\method extracts an effective feature vector of a small size from a subsequence by exploiting the characteristics of workload sequences.
Based on the feature vectors, \method accurately detects test samples of the unknown class by  constructing SVD-based detectors for each class.
Experiments show that \method outperforms existing open-set recognition methods, simultaneously achieving higher performance for known classes and the unknown class.
\green{Future works include a novel feature extraction method considering the association between heterogeneous fields, and extending the proposed method for other applications such as malware detection.}

\begin{acks}
This work is supported by Samsung Electronics Co., Ltd. 
The Institute of Engineering Research and ICT at Seoul National University provided research facilities for this work. 
U Kang is the corresponding author.
\end{acks}

\bibliographystyle{ACM-Reference-Format}
\bibliography{mybib}


\begin{thebibliography}{30}


\ifx \showCODEN    \undefined \def \showCODEN     #1{\unskip}     \fi
\ifx \showDOI      \undefined \def \showDOI       #1{#1}\fi
\ifx \showISBNx    \undefined \def \showISBNx     #1{\unskip}     \fi
\ifx \showISBNxiii \undefined \def \showISBNxiii  #1{\unskip}     \fi
\ifx \showISSN     \undefined \def \showISSN      #1{\unskip}     \fi
\ifx \showLCCN     \undefined \def \showLCCN      #1{\unskip}     \fi
\ifx \shownote     \undefined \def \shownote      #1{#1}          \fi
\ifx \showarticletitle \undefined \def \showarticletitle #1{#1}   \fi
\ifx \showURL      \undefined \def \showURL       {\relax}        \fi
\providecommand\bibfield[2]{#2}
\providecommand\bibinfo[2]{#2}
\providecommand\natexlab[1]{#1}
\providecommand\showeprint[2][]{arXiv:#2}

\bibitem[\protect\citeauthoryear{??}{ddr}{2013}]%
        {ddr4}
 \bibinfo{year}{2013}\natexlab{}.
\newblock \showarticletitle{SDRAM STANDARD}.
\newblock  (\bibinfo{year}{2013}).
\newblock
\urldef\tempurl%
\url{https://www.jedec.org/sites/default/files/docs/JESD79-4.pdf}
\showURL{%
\tempurl}


\bibitem[\protect\citeauthoryear{Ali, Shiaeles, Bendiab, and Ghita}{Ali
  et~al\mbox{.}}{2020}]%
        {Ali20}
\bibfield{author}{\bibinfo{person}{Muhammad Ali}, \bibinfo{person}{Stavros
  Shiaeles}, \bibinfo{person}{Gueltoum Bendiab}, {and} \bibinfo{person}{Bogdan
  Ghita}.} \bibinfo{year}{2020}\natexlab{}.
\newblock \showarticletitle{MALGRA: Machine learning and N-gram malware feature
  extraction and detection system}.
\newblock \bibinfo{journal}{\emph{Electronics}} \bibinfo{volume}{9},
  \bibinfo{number}{11} (\bibinfo{year}{2020}), \bibinfo{pages}{1777}.
\newblock


\bibitem[\protect\citeauthoryear{Bendale and Boult}{Bendale and Boult}{2016}]%
        {BendaleB16}
\bibfield{author}{\bibinfo{person}{Abhijit Bendale} {and}
  \bibinfo{person}{Terrance~E. Boult}.} \bibinfo{year}{2016}\natexlab{}.
\newblock \showarticletitle{Towards Open Set Deep Networks}. In
  \bibinfo{booktitle}{\emph{CVPR}}. \bibinfo{publisher}{{IEEE} Computer
  Society}, \bibinfo{pages}{1563--1572}.
\newblock


\bibitem[\protect\citeauthoryear{Chen, Yao, Zhang, Wang, Chang, and Nie}{Chen
  et~al\mbox{.}}{2020}]%
        {ChenYZWCN20}
\bibfield{author}{\bibinfo{person}{Kaixuan Chen}, \bibinfo{person}{Lina Yao},
  \bibinfo{person}{Dalin Zhang}, \bibinfo{person}{Xianzhi Wang},
  \bibinfo{person}{Xiaojun Chang}, {and} \bibinfo{person}{Feiping Nie}.}
  \bibinfo{year}{2020}\natexlab{}.
\newblock \showarticletitle{A Semisupervised Recurrent Convolutional Attention
  Model for Human Activity Recognition}.
\newblock \bibinfo{journal}{\emph{{IEEE} Trans. Neural Networks Learn. Syst.}}
  \bibinfo{volume}{31}, \bibinfo{number}{5} (\bibinfo{year}{2020}),
  \bibinfo{pages}{1747--1756}.
\newblock


\bibitem[\protect\citeauthoryear{Hassen and Chan}{Hassen and Chan}{2020}]%
        {HassenC20}
\bibfield{author}{\bibinfo{person}{Mehadi Hassen} {and}
  \bibinfo{person}{Philip~K. Chan}.} \bibinfo{year}{2020}\natexlab{}.
\newblock \showarticletitle{Learning a Neural-network-based Representation for
  Open Set Recognition}. In \bibinfo{booktitle}{\emph{SDM}}.
  \bibinfo{publisher}{{SIAM}}, \bibinfo{pages}{154--162}.
\newblock


\bibitem[\protect\citeauthoryear{Hendrycks and Gimpel}{Hendrycks and
  Gimpel}{2017}]%
        {HendrycksG17}
\bibfield{author}{\bibinfo{person}{Dan Hendrycks} {and} \bibinfo{person}{Kevin
  Gimpel}.} \bibinfo{year}{2017}\natexlab{}.
\newblock \showarticletitle{A Baseline for Detecting Misclassified and
  Out-of-Distribution Examples in Neural Networks}. In
  \bibinfo{booktitle}{\emph{5th International Conference on Learning
  Representations, {ICLR} 2017, Toulon, France, April 24-26, 2017, Conference
  Track Proceedings}}. \bibinfo{publisher}{OpenReview.net}.
\newblock


\bibitem[\protect\citeauthoryear{Jang, Choi, Jung, and Kang}{Jang
  et~al\mbox{.}}{2018}]%
        {JangCJK18}
\bibfield{author}{\bibinfo{person}{Jun{-}Gi Jang}, \bibinfo{person}{Dongjin
  Choi}, \bibinfo{person}{Jinhong Jung}, {and} \bibinfo{person}{U Kang}.}
  \bibinfo{year}{2018}\natexlab{}.
\newblock \showarticletitle{Zoom-SVD: Fast and Memory Efficient Method for
  Extracting Key Patterns in an Arbitrary Time Range}. In
  \bibinfo{booktitle}{\emph{CIKM}}. \bibinfo{publisher}{{ACM}},
  \bibinfo{pages}{1083--1092}.
\newblock


\bibitem[\protect\citeauthoryear{Jang and Kang}{Jang and Kang}{2020}]%
        {JangK20}
\bibfield{author}{\bibinfo{person}{Jun{-}Gi Jang} {and} \bibinfo{person}{U
  Kang}.} \bibinfo{year}{2020}\natexlab{}.
\newblock \showarticletitle{D-Tucker: Fast and Memory-Efficient Tucker
  Decomposition for Dense Tensors}. In \bibinfo{booktitle}{\emph{ICDE}}.
  \bibinfo{publisher}{{IEEE}}, \bibinfo{pages}{1850--1853}.
\newblock


\bibitem[\protect\citeauthoryear{Jolliffe}{Jolliffe}{2002}]%
        {jolliffe2002principal}
\bibfield{author}{\bibinfo{person}{Ian Jolliffe}.}
  \bibinfo{year}{2002}\natexlab{}.
\newblock \bibinfo{booktitle}{\emph{Principal component analysis}}.
\newblock \bibinfo{publisher}{Wiley Online Library}.
\newblock


\bibitem[\protect\citeauthoryear{Lee, Yu, and Yu}{Lee et~al\mbox{.}}{2020}]%
        {LeeYY20}
\bibfield{author}{\bibinfo{person}{Dongha Lee}, \bibinfo{person}{Sehun Yu},
  {and} \bibinfo{person}{Hwanjo Yu}.} \bibinfo{year}{2020}\natexlab{}.
\newblock \showarticletitle{Multi-Class Data Description for
  Out-of-distribution Detection}. In \bibinfo{booktitle}{\emph{KDD}}.
  \bibinfo{publisher}{{ACM}}, \bibinfo{pages}{1362--1370}.
\newblock


\bibitem[\protect\citeauthoryear{Lee, Lee, Lee, and Shin}{Lee
  et~al\mbox{.}}{2018}]%
        {LeeLLS18}
\bibfield{author}{\bibinfo{person}{Kimin Lee}, \bibinfo{person}{Kibok Lee},
  \bibinfo{person}{Honglak Lee}, {and} \bibinfo{person}{Jinwoo Shin}.}
  \bibinfo{year}{2018}\natexlab{}.
\newblock \showarticletitle{A Simple Unified Framework for Detecting
  Out-of-Distribution Samples and Adversarial Attacks}. In
  \bibinfo{booktitle}{\emph{NeurIPS 2018}}. \bibinfo{pages}{7167--7177}.
\newblock


\bibitem[\protect\citeauthoryear{Li, Yao, Chang, Zhan, Sun, and Zhang}{Li
  et~al\mbox{.}}{2019}]%
        {LiYCZSZ19}
\bibfield{author}{\bibinfo{person}{Zhihui Li}, \bibinfo{person}{Lina Yao},
  \bibinfo{person}{Xiaojun Chang}, \bibinfo{person}{Kun Zhan},
  \bibinfo{person}{Jiande Sun}, {and} \bibinfo{person}{Huaxiang Zhang}.}
  \bibinfo{year}{2019}\natexlab{}.
\newblock \showarticletitle{Zero-shot event detection via event-adaptive
  concept relevance mining}.
\newblock \bibinfo{journal}{\emph{Pattern Recognit.}}  \bibinfo{volume}{88}
  (\bibinfo{year}{2019}), \bibinfo{pages}{595--603}.
\newblock


\bibitem[\protect\citeauthoryear{Liang, Li, and Srikant}{Liang
  et~al\mbox{.}}{2018}]%
        {LiangLS18}
\bibfield{author}{\bibinfo{person}{Shiyu Liang}, \bibinfo{person}{Yixuan Li},
  {and} \bibinfo{person}{R. Srikant}.} \bibinfo{year}{2018}\natexlab{}.
\newblock \showarticletitle{Enhancing The Reliability of Out-of-distribution
  Image Detection in Neural Networks}. In \bibinfo{booktitle}{\emph{ICLR}}.
  \bibinfo{publisher}{OpenReview.net}.
\newblock


\bibitem[\protect\citeauthoryear{Liu, Wang, Owens, and Li}{Liu
  et~al\mbox{.}}{2020}]%
        {LiuWOL20}
\bibfield{author}{\bibinfo{person}{Weitang Liu}, \bibinfo{person}{Xiaoyun
  Wang}, \bibinfo{person}{John~D. Owens}, {and} \bibinfo{person}{Yixuan Li}.}
  \bibinfo{year}{2020}\natexlab{}.
\newblock \showarticletitle{Energy-based Out-of-distribution Detection}. In
  \bibinfo{booktitle}{\emph{NeurIPS}}.
\newblock


\bibitem[\protect\citeauthoryear{Luo, Chang, Nie, Yang, Hauptmann, and
  Zheng}{Luo et~al\mbox{.}}{2018}]%
        {LuoCNYHZ18}
\bibfield{author}{\bibinfo{person}{Minnan Luo}, \bibinfo{person}{Xiaojun
  Chang}, \bibinfo{person}{Liqiang Nie}, \bibinfo{person}{Yi Yang},
  \bibinfo{person}{Alexander~G. Hauptmann}, {and} \bibinfo{person}{Qinghua
  Zheng}.} \bibinfo{year}{2018}\natexlab{}.
\newblock \showarticletitle{An Adaptive Semisupervised Feature Analysis for
  Video Semantic Recognition}.
\newblock \bibinfo{journal}{\emph{{IEEE} Trans. Cybern.}} \bibinfo{volume}{48},
  \bibinfo{number}{2} (\bibinfo{year}{2018}), \bibinfo{pages}{648--660}.
\newblock


\bibitem[\protect\citeauthoryear{Mutinda, Mwangi, and Okeyo}{Mutinda
  et~al\mbox{.}}{2021}]%
        {Mutinda21}
\bibfield{author}{\bibinfo{person}{James Mutinda}, \bibinfo{person}{Waweru
  Mwangi}, {and} \bibinfo{person}{George Okeyo}.}
  \bibinfo{year}{2021}\natexlab{}.
\newblock \showarticletitle{Lexicon-pointed hybrid N-gram Features Extraction
  Model (LeNFEM) for sentence level sentiment analysis}.
\newblock \bibinfo{journal}{\emph{Engineering Reports}} \bibinfo{volume}{3},
  \bibinfo{number}{8} (\bibinfo{year}{2021}), \bibinfo{pages}{e12374}.
\newblock


\bibitem[\protect\citeauthoryear{Osi{\'n}ski, Stefanowski, and
  Weiss}{Osi{\'n}ski et~al\mbox{.}}{2004}]%
        {osinski2004lingo}
\bibfield{author}{\bibinfo{person}{Stanis{\l}aw Osi{\'n}ski},
  \bibinfo{person}{Jerzy Stefanowski}, {and} \bibinfo{person}{Dawid Weiss}.}
  \bibinfo{year}{2004}\natexlab{}.
\newblock \showarticletitle{Lingo: Search results clustering algorithm based on
  singular value decomposition}.
\newblock In \bibinfo{booktitle}{\emph{Intelligent information processing and
  web mining}}. \bibinfo{publisher}{Springer}, \bibinfo{pages}{359--368}.
\newblock


\bibitem[\protect\citeauthoryear{Oza and Patel}{Oza and Patel}{2019}]%
        {OzaP19}
\bibfield{author}{\bibinfo{person}{Poojan Oza} {and} \bibinfo{person}{Vishal~M.
  Patel}.} \bibinfo{year}{2019}\natexlab{}.
\newblock \showarticletitle{{C2AE:} Class Conditioned Auto-Encoder for Open-Set
  Recognition}. In \bibinfo{booktitle}{\emph{CVPR}}.
  \bibinfo{publisher}{Computer Vision Foundation / {IEEE}},
  \bibinfo{pages}{2307--2316}.
\newblock


\bibitem[\protect\citeauthoryear{Scheirer, de~Rezende~Rocha, Sapkota, and
  Boult}{Scheirer et~al\mbox{.}}{2013}]%
        {ScheirerRSB13}
\bibfield{author}{\bibinfo{person}{Walter~J. Scheirer},
  \bibinfo{person}{Anderson de Rezende~Rocha}, \bibinfo{person}{Archana
  Sapkota}, {and} \bibinfo{person}{Terrance~E. Boult}.}
  \bibinfo{year}{2013}\natexlab{}.
\newblock \showarticletitle{Toward Open Set Recognition}.
\newblock \bibinfo{journal}{\emph{{IEEE} Trans. Pattern Anal. Mach. Intell.}}
  \bibinfo{volume}{35}, \bibinfo{number}{7} (\bibinfo{year}{2013}),
  \bibinfo{pages}{1757--1772}.
\newblock


\bibitem[\protect\citeauthoryear{Shu, Xu, and Liu}{Shu et~al\mbox{.}}{2017}]%
        {ShuXL17}
\bibfield{author}{\bibinfo{person}{Lei Shu}, \bibinfo{person}{Hu Xu}, {and}
  \bibinfo{person}{Bing Liu}.} \bibinfo{year}{2017}\natexlab{}.
\newblock \showarticletitle{{DOC:} Deep Open Classification of Text Documents}.
  In \bibinfo{booktitle}{\emph{EMNLP}}. \bibinfo{publisher}{Association for
  Computational Linguistics}, \bibinfo{pages}{2911--2916}.
\newblock


\bibitem[\protect\citeauthoryear{Simek, Fujarewicz, {\'S}wierniak, Kimmel,
  Jarzab, Wiench, and Rzeszowska}{Simek et~al\mbox{.}}{2004}]%
        {simek2004using}
\bibfield{author}{\bibinfo{person}{Krzysztof Simek}, \bibinfo{person}{Krzysztof
  Fujarewicz}, \bibinfo{person}{Andrzej {\'S}wierniak}, \bibinfo{person}{Marek
  Kimmel}, \bibinfo{person}{Barbara Jarzab}, \bibinfo{person}{Ma{\l}gorzata
  Wiench}, {and} \bibinfo{person}{Joanna Rzeszowska}.}
  \bibinfo{year}{2004}\natexlab{}.
\newblock \showarticletitle{Using SVD and SVM methods for selection,
  classification, clustering and modeling of DNA microarray data}.
\newblock \bibinfo{journal}{\emph{Engineering Applications of Artificial
  Intelligence}} \bibinfo{volume}{17}, \bibinfo{number}{4}
  (\bibinfo{year}{2004}), \bibinfo{pages}{417--427}.
\newblock


\bibitem[\protect\citeauthoryear{Sun, Yang, Zhang, Ling, and Peng}{Sun
  et~al\mbox{.}}{2020}]%
        {SunYZLP20}
\bibfield{author}{\bibinfo{person}{Xin Sun}, \bibinfo{person}{Zhenning Yang},
  \bibinfo{person}{Chi Zhang}, \bibinfo{person}{Keck~Voon Ling}, {and}
  \bibinfo{person}{Guohao Peng}.} \bibinfo{year}{2020}\natexlab{}.
\newblock \showarticletitle{Conditional Gaussian Distribution Learning for Open
  Set Recognition}. In \bibinfo{booktitle}{\emph{CVPR}}.
  \bibinfo{publisher}{{IEEE}}, \bibinfo{pages}{13477--13486}.
\newblock


\bibitem[\protect\citeauthoryear{Sun, Guo, and Li}{Sun et~al\mbox{.}}{2021}]%
        {Sun21}
\bibfield{author}{\bibinfo{person}{Yiyou Sun}, \bibinfo{person}{Chuan Guo},
  {and} \bibinfo{person}{Yixuan Li}.} \bibinfo{year}{2021}\natexlab{}.
\newblock \showarticletitle{ReAct: Out-of-distribution Detection With Rectified
  Activations}.
\newblock \bibinfo{journal}{\emph{Advances in Neural Information Processing
  Systems}} (\bibinfo{year}{2021}).
\newblock


\bibitem[\protect\citeauthoryear{Tomovic, Janicic, and Keselj}{Tomovic
  et~al\mbox{.}}{2006}]%
        {TomovicJK06}
\bibfield{author}{\bibinfo{person}{Andrija Tomovic}, \bibinfo{person}{Predrag
  Janicic}, {and} \bibinfo{person}{Vlado Keselj}.}
  \bibinfo{year}{2006}\natexlab{}.
\newblock \showarticletitle{n-Gram-based classification and unsupervised
  hierarchical clustering of genome sequences}.
\newblock \bibinfo{journal}{\emph{Comput. Methods Programs Biomed.}}
  \bibinfo{volume}{81}, \bibinfo{number}{2} (\bibinfo{year}{2006}),
  \bibinfo{pages}{137--153}.
\newblock


\bibitem[\protect\citeauthoryear{Wall, Rechtsteiner, and Rocha}{Wall
  et~al\mbox{.}}{2003}]%
        {wall2003singular}
\bibfield{author}{\bibinfo{person}{Michael~E Wall}, \bibinfo{person}{Andreas
  Rechtsteiner}, {and} \bibinfo{person}{Luis~M Rocha}.}
  \bibinfo{year}{2003}\natexlab{}.
\newblock \showarticletitle{Singular value decomposition and principal
  component analysis}.
\newblock In \bibinfo{booktitle}{\emph{A practical approach to microarray data
  analysis}}. \bibinfo{publisher}{Springer}, \bibinfo{pages}{91--109}.
\newblock


\bibitem[\protect\citeauthoryear{Yoo, Cho, Kim, and Kang}{Yoo
  et~al\mbox{.}}{2019a}]%
        {YooCKK19}
\bibfield{author}{\bibinfo{person}{Jaemin Yoo}, \bibinfo{person}{Minyong Cho},
  \bibinfo{person}{Taebum Kim}, {and} \bibinfo{person}{U Kang}.}
  \bibinfo{year}{2019}\natexlab{a}.
\newblock \showarticletitle{Knowledge Extraction with No Observable Data}. In
  \bibinfo{booktitle}{\emph{NeurIPS}}. \bibinfo{pages}{2701--2710}.
\newblock


\bibitem[\protect\citeauthoryear{Yoo, Jeon, and Kang}{Yoo
  et~al\mbox{.}}{2019b}]%
        {YooJK19}
\bibfield{author}{\bibinfo{person}{Jaemin Yoo}, \bibinfo{person}{Hyunsik Jeon},
  {and} \bibinfo{person}{U Kang}.} \bibinfo{year}{2019}\natexlab{b}.
\newblock \showarticletitle{Belief Propagation Network for Hard Inductive
  Semi-Supervised Learning}. In \bibinfo{booktitle}{\emph{IJCAI}},
  \bibfield{editor}{\bibinfo{person}{Sarit Kraus}} (Ed.).
  \bibinfo{publisher}{ijcai.org}, \bibinfo{pages}{4178--4184}.
\newblock


\bibitem[\protect\citeauthoryear{Yoshihashi, Shao, Kawakami, You, Iida, and
  Naemura}{Yoshihashi et~al\mbox{.}}{2019}]%
        {YoshihashiSKYIN19}
\bibfield{author}{\bibinfo{person}{Ryota Yoshihashi}, \bibinfo{person}{Wen
  Shao}, \bibinfo{person}{Rei Kawakami}, \bibinfo{person}{Shaodi You},
  \bibinfo{person}{Makoto Iida}, {and} \bibinfo{person}{Takeshi Naemura}.}
  \bibinfo{year}{2019}\natexlab{}.
\newblock \showarticletitle{Classification-Reconstruction Learning for Open-Set
  Recognition}. In \bibinfo{booktitle}{\emph{CVPR}}.
  \bibinfo{publisher}{Computer Vision Foundation / {IEEE}},
  \bibinfo{pages}{4016--4025}.
\newblock


\bibitem[\protect\citeauthoryear{Zhang, Xiao, Mercaldo, Ni, Martinelli, and
  Sangaiah}{Zhang et~al\mbox{.}}{2019}]%
        {ZhangXMNMS19}
\bibfield{author}{\bibinfo{person}{Hanqi Zhang}, \bibinfo{person}{Xi Xiao},
  \bibinfo{person}{Francesco Mercaldo}, \bibinfo{person}{Shiguang Ni},
  \bibinfo{person}{Fabio Martinelli}, {and} \bibinfo{person}{Arun~Kumar
  Sangaiah}.} \bibinfo{year}{2019}\natexlab{}.
\newblock \showarticletitle{Classification of ransomware families with machine
  learning based on N-gram of opcodes}.
\newblock \bibinfo{journal}{\emph{Future Gener. Comput. Syst.}}
  \bibinfo{volume}{90} (\bibinfo{year}{2019}), \bibinfo{pages}{211--221}.
\newblock


\bibitem[\protect\citeauthoryear{Zhang, Srivastava, Nori, Kannan, and
  Prasanna}{Zhang et~al\mbox{.}}{2022}]%
        {ZhangSNKP22}
\bibfield{author}{\bibinfo{person}{Pengmiao Zhang}, \bibinfo{person}{Ajitesh
  Srivastava}, \bibinfo{person}{Anant~V. Nori}, \bibinfo{person}{Rajgopal
  Kannan}, {and} \bibinfo{person}{Viktor~K. Prasanna}.}
  \bibinfo{year}{2022}\natexlab{}.
\newblock \showarticletitle{Fine-grained address segmentation for
  attention-based variable-degree prefetching}. In
  \bibinfo{booktitle}{\emph{{CF} '22: 19th {ACM} International Conference on
  Computing Frontiers, Turin, Italy, May 17 - 22, 2022}}.
  \bibinfo{publisher}{{ACM}}, \bibinfo{pages}{103--112}.
\newblock


\end{thebibliography}

\end{document}